\documentclass[journal]{IEEEtran}

\usepackage{ifpdf}
\usepackage{cite}
\ifCLASSINFOpdf
  \usepackage[pdftex]{graphicx}

\else

\fi

\usepackage{amsmath}

\usepackage{algorithmic}

\usepackage{array}

\usepackage{fixltx2e}

\usepackage{float}
\usepackage{stfloats}

\ifCLASSOPTIONcaptionsoff
 \usepackage[nomarkers]{endfloat}
 \let\MYoriglatexcaption\caption
 \renewcommand{\caption}[2][\relax]{\MYoriglatexcaption[#2]{#2}}
\fi

\usepackage{url}

\usepackage{tabularx}
\newcolumntype{L}[1]{>{\raggedright\let\newline\\\arraybackslash\hspace{0pt}}m{#1}}
\newcolumntype{C}[1]{>{\centering\let\newline\\\arraybackslash\hspace{0pt}}m{#1}}
\newcolumntype{R}[1]{>{\raggedleft\let\newline\\\arraybackslash\hspace{0pt}}m{#1}}
\usepackage{multirow}
\usepackage[export]{adjustbox}

\usepackage{subcaption}
\usepackage[utf8]{inputenc}
\usepackage{booktabs,subcaption,amsfonts,dcolumn}

\usepackage{multirow}
\usepackage{chngpage}
\usepackage{adjustbox}
\usepackage[export]{adjustbox}
\usepackage{slashbox}
\usepackage{float}
\usepackage{comment}
\usepackage{siunitx}

\usepackage[breaklinks]{hyperref}
\hypersetup{
    colorlinks=true,
    linkcolor=blue,
    filecolor=magenta,
    urlcolor=cyan,
}

\usepackage{lipsum}
\usepackage[mathlines]{lineno}

\begin{document}

%
\title{Evaluation of Sampling Methods for Robotic Sediment Sampling Systems}
%
%
%


\author{Jun~Han~Bae,
        Wonse~Jo,
        Jee~Hwan~Park,
        Richard~M.~Voyles, \textit{Senior Member, IEEE},
        Sara~K.~McMillan,
        and~Byung-Cheol~Min$^*$, \textit{Member, IEEE}
\thanks{Manuscript accepted May 26, 2020.

J.H. Bae, W. Jo, and B.-C. Min are with SMART Lab, Department of Computer and Information Technology, Purdue University, West Lafayette, IN 47907, USA.

J.H. Park is with SMART Lab, Department of Computer and Information Technology, and with the School of Mechanical Engineering, Purdue University, West Lafayette, IN 47907, USA.

R. M. Voyles is with the Collaborative Robotics Lab, School of Engineering Technology, Purdue University, West Lafayette, IN 47907, USA.

S. K. McMillan is with the Department of Agricultural \& Biological Engineering, Purdue University, West Lafayette, IN 47907, USA.

$^*$ Corresponding author email: {\tt minb@purdue.edu}.}}

%
%

\markboth{Preprint version accepted to IEEE JOURNAL OF OCEANIC ENGINEERING}%
{\MakeLowercase{\textit{Bae and Min}}:}
%



\maketitle

\begin{abstract}
Analysis of sediments from rivers, lakes, reservoirs, wetlands and other constructed surface water impoundments is an important tool to characterize the function and health of these systems, but is generally carried out manually. This is costly and can be hazardous and difficult for humans due to inaccessibility, contamination, or availability of required equipment. Robotic sampling systems can ease these burdens, but little work has examined the efficiency of such sampling means and no prior work has investigated the quality of the resulting samples. This paper presents an experimental study that evaluates and optimizes sediment sampling patterns applied to a robot sediment sampling system that allows collection of minimally-disturbed sediment cores from natural and man-made water bodies for various sediment types. To meet this need, we developed and tested a robotic sampling platform in the laboratory to test functionality under a range of sediment types and operating conditions. Specifically, we focused on three patterns by which a cylindrical coring device was driven into the sediment (linear, helical, and zig-zag) for three sediment types (coarse sand, medium sand, and silt). The results show that the optimal sampling pattern varies depending on the type of sediment and can be optimized based on the sampling objective. We examined two sampling objectives: maximizing the mass of minimally disturbed sediment and minimizing the power per mass of sample. This study provides valuable data to aid in the selection of optimal sediment coring methods for various applications and builds a solid foundation for future field testing under a range of environmental conditions.
\end{abstract}


\IEEEpeerreviewmaketitle

\section{Introduction}
\label{Section1}
Sediment consists of solid particles of mineral and organic material that can be transported by water and eventually deposited on the bed or bottom of a body of water \cite{international2011sediment}. Sediment is an important component in the natural geochemical cycle and moves from land to ocean by river systems and vice versa \cite{international2011sediment}. Sediments in surface water systems are intrinsically linked to conditions in ecological (e.g., food webs) and geochemical (e.g., water quality) contexts which require accurate, reproducible, and representative sampling of their respective subsurface systems. Traditional methods of collection via wading, boating, or with diver assistance can be difficult, costly, and hazardous to humans due to inaccessibility, contamination, or lack of availability of equipment \cite{sekellick_water_2013,noauthor_sediment_nodate}. Sampling in contaminated areas or in rivers with rapidly flowing or deep water (relative to humans) pose significant and unique dangers to workers completing these tasks. Lastly, inconsistent topography and bathymetry can make desired sampling schemes difficult to implement resulting in data that is not fully representative because certain areas are over or under sampled. 

Recent advances in robotic technology have demonstrated the potential to fill this gap and offer solutions that can be both economical, effective and safe in these environments. For example, a few prior robots have been developed to collect sediment in oceanic environments \cite{paull_deep_2001,yoshida_two-stage_2006,yoshida_deepest_2007,whitcomb_navigation_2010,sathianarayanan_deep_2013}. These robots are large and heavy due to typical conditions in the open ocean, so they are not practical for smaller water basins like rivers, streams, lakes, or reservoirs, but they have proven effective in removing humans from the immediate dangers of deep-water sampling. Remotely Operated underwater Vehicles (ROVs) for smaller bodies of water have been explored that are capable of sampling sediment to solve the various challenges mentioned above \cite{gao2015development,sakagami_development_2013,yokoi_improvement_2014}. However, these research systems are still under development with the primary focus on navigation, path planning, and orientation control of the underwater vehicle with less consideration of the impact of sampling technique on the integrity of the sample collected.

Understanding the needs of sediment researchers, the characteristics of the sediment itself, and the physical constraints of the body of water are imperative for developing a robust robotic sediment sampling system. A clear understanding of how the specific sediment type interacts with the sampling equipment becomes more critical when sampling with robotic systems as compared to manual sampling with human oversight and adaptation. Traditional methods of manual sampling, particularly those in shallow water with the wading method, allow the worker to change approaches rapidly based on sediment compaction, presence of rocks or plant roots, or simply different soil textures. However, with robotic samplers, these conditions need to be assessed in real time and the penetration force adjusted to sample to the desired depth. This requires analysis of the resistance force, which strongly influences the stability, mass, and portability of the sampling platform. In addition, physical disturbances to the sample during the sampling process can affect the physical, geochemical, and biological conditions of the sample and should be minimized. Thus, the choice of sampling pattern is both critical and challenging.

To meet these challenges, we developed a robotic sediment sampling platform for man-made water bodies and tested it in the laboratory under simulated environmental conditions to find the optimal sediment sampling method based on the type of sediment. We selected three sediment types (coarse sand, medium sand, and silt) commonly found in reservoirs, lakes, and rivers in temperate climates and applied three sampling patterns under two optimization objectives: sampled mass and work efficiency. We measured the mass of the sediment sample and work efficiency for the purpose of evaluating the performance of the proposed sampling system. Also, we assumed that the greater amount of the collected sample would mean the higher quality sample as they could provide more data for the analysis. We only tested one commercially-available coring tool with fixed diameter, cutting shoe angle, and wall thickness, but we chose a diameter relevant to immediate needs for our early prototypes. The main contributions of this paper are to: 1) provide a thorough analysis of our proposed approach based on the robotic technology that will improve reproducible capability and minimize risks associated with manual sampling, and 2) identify optimal coring techniques that can be applied under multiple underwater sediment sampling applications.

The paper is organized as follows: Section~\ref{sec:background} presents the background and highlights the constraints of current sediment sampling technologies. In Section~\ref{sec:overview}, we provide an overview of the developed sediment sampling platform.  We also describe our mathematical analysis of the sampling patterns and the three types of sediment used in testing based on particle distribution. We describe our methodology in Section~\ref{sec:methodology} based on multi-objective optimization. In Section~\ref{sec:experiment}, we explain the experimental procedure, statistical approach, and results of our tests. We present an analysis of the data based on these measurements in Section~\ref{sec:analysis}. We conclude the paper and discuss future work in Section~\ref{sec:conclusion}.

\section{Background}
\label{sec:background}
\subsection{Sediment Sampling Methods and Equipment}
Multiple types of physical, chemical, and biological analyses of surface water systems require sampling of subsurface sediments. Defining the purpose of sampling is essential to determining the best parameter values such as the number of samples and locations collected, depth of sample required, and necessity of maintaining the stratifications of the sample with depth. Some examples include food web bioassays, pollutant discharge monitoring, contaminant source identification, trace elements analysis, and contribution of sediment processes to surface water bio-geochemistry such as eutrophication \cite{kasich2012sediment}\cite{shelton1994guidelines}.

Selecting the appropriate sampling instrument and method depends on the sampling objective, type of sediment, and sampling location (i.e., water depth) \cite{noauthor_lake_1997,iaea_collection_2003,kasich2012sediment}. Based on previous studies \cite{kasich2012sediment,skilbeck_sediment_2017,us_epa_methods_2015}, sediment samplers can be classified into four basic categories: auger, dredge, grab, and core sampler. The first three samplers are appropriate for a composite sampling of the sediment surface, typically within 20-30 cm of the boundary of the overlying water column \cite{skilbeck_sediment_2017}. The auger is used to sample the surface sediment of sub-aerial environments. The dredge and grab samplers are preferred for underwater needs in collecting consolidated sediment with coarser particle size, collecting large volumes of sediment, and surveying large areas. In all three cases, the sediment is highly disturbed and eliminates analysis of any vertical heterogeneity.

Conversely, core samplers are widely used to maintain in-situ sediment layers in a wide range of sediment types (e.g., unconsolidated to semi-consolidated sediments; fine to medium particle size sediments). They can extend deep into the sediment ($>$ 3 m in some cases), characterize sediments, and observe the historical changes preserved in the vertical distribution of the sediment \cite{noauthor_sediment_nodate}. They are also frequently used to maintain the distribution of electron acceptors for microbial communities when assessing the impact of sediment processes on water quality \cite{BryantLeeD.2012Rosm}. The specific type of core samplers can be selected based on the water depth, core size, and sediment type \cite{noauthor_standard_2014_2}. The typical core sampling methods include gravity coring, box coring, piston coring, and vibro-coring, which make use of free fall (weight), piston, or hydraulic energy \cite{skilbeck_sediment_2017,us_epa_methods_2015}. 

\begin{figure}[t!] 
  \centering
  \includegraphics[width=0.28\textwidth]{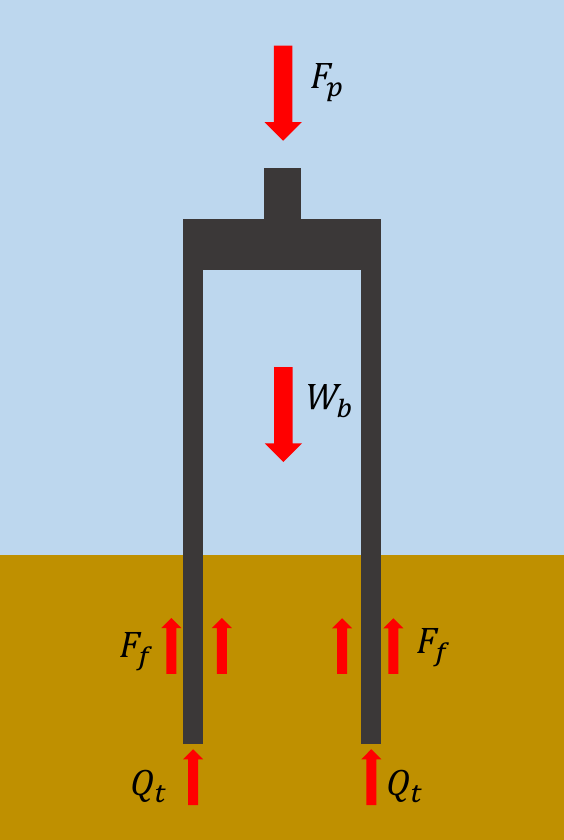}  
  \caption{Forces acting on the sampler \cite{rocker1985handbook}; $F_{p}$ is the penetration force, $F_{e}$ is the external driven force, $W_{b}$ is the buoyant weight of the sampler, $Q_{t}$ is the bearing capacity at the tip, and $F_{t}$ is the side friction. }
  \label{forcediagram} 
\end{figure}

The forces acting on the core sampler can be simplified as illustrated in Fig. \ref{forcediagram}, and the penetration force is given
\begin{equation}
\label{Equationfs}
F_{p}=F_{e}+W_{b}-Q_{t}-F_{t}
\end{equation}

\noindent where $F_{p}$ is the penetration force, $F_{e}$ is the external driven force, $W_{b}$ is the buoyant weight of the sampler, $Q_{t}$ is the bearing capacity at the tip, and $F_{t}$ is the side friction \cite{rocker1985handbook}. When $F_{p}>0$, the penetration with the sampling core is successful. This means that a greater force with $F_{e}$ and $W_{b}$ should be generated than the resistance force $Q_{t}+F_{t}$, i.e., $F_{e}+W_{b}>Q_{t}+F_{t}$, to sample the sediment. Finding an optimal penetration force is needed to develop an efficient robotic sediment sampling system. While it is possible to maximize the penetration force by increasing the external force and weight of the platform, it is inefficient from an optimal robot design and control standpoint. Also, the resistance force heavily depends on the type of the sediment, and predicting it is challenging due to uncertainties such as sediment texture, bed slope, or presence of unexpected obstacles (e.g., plant roots). Measuring sediment characteristics is feasible for manual and ship-based sampling, but we defer this discussion, as small robots have limited resources and cannot carry multiple payloads.

\subsection{State of the Art Robotic Technology}
Unmanned platforms such as ROVs and Unmanned Ground Vehicles (UGVs) have been utilized in various studies to reduce the processing time and associated risks of manual sampling by automating transport and control. For example, a tethered underwater vehicle with a detachable core sampler was used to sample sediment with the help of a boat on the surface, as described in \cite{gao2015development}. This system requires a crane to lower the ROV and an anchor to the bottom. The ROV has fin-based thrusters, but is tethered to the anchor as well as the human operators on the ship at the surface. Humans remotely select a point to sample and drop the negatively buoyant sampler from the ROV into the soil, then hoist everything up with the crane. A smaller operator-portable ROV with sample coring cylinders was introduced in \cite{sakagami_development_2013,yokoi_improvement_2014}. At 2.3 m in length and 34 kg in mass (in the air), it is more portable than previous platforms. Vertically directed thrusters, rather than negative buoyancy, generate forces that allow the core to penetrate the sediment. However, due to the limitations of the electric thrusters, the penetration depth of the sediment in field experiments was limited to approximately 0.16 m \cite{sakagami_development_2013}. Another study explored cooperation between human operators and robotic teams for sampling along the shoreline \cite{deusdado_sediment_2016}. For this system, an operator decides which specific area to sample based on images from a high-resolution camera (similar to \cite{gao2015development}, above), attached to an Unmanned Aerial Vehicle (UAV) instead of an ROV. A UGV then travels to the desired location and samples the sediment. 

The resistance force $Q_{t}+F_{t}$ due to the bearing capacity while sampling can cause instability of the sampling platform when the platform is unable to generate the penetration force, i.e., $F_{e}+W_{b} < Q_{t}+F_{t}$. In the case of \cite{bae_tri-sedimentbot:_2016}, it is crucial to maintain the stability of the underwater sediment sampler during the sampling process by countering the linear and rotational motion of the drill \cite{bae_tri-sedimentbot:_2016}. A feedback controller was implemented to control stability by measuring the difference between the reference and actual orientations. A closed-loop control system has shown better performance compared to an open-loop control system, including a shorter period of sampling time and better stability, based on the weight of sediments \cite{bae_tri-sedimentbot:_2016}.
 Those studies demonstrate that an understanding of the sampling environment and material properties is essential to the design and development of an automated sampling system. In the case of underwater sediment sampling, however, few research studies have been published \cite{escudero_experimental_2017}. To develop a more effective automated sediment sampling system, a thorough understanding of the underwater environment\cite{Ocean_Engineering_Journal-03,Ocean_Engineering_Journal-04} and sediment sampling \cite{Ocean_Engineering_Journal-01,Ocean_Engineering_Journal-02} is necessary, particularly in more challenging environments.

\section{Sediment Sampling Platform Development}
\label{sec:overview}
\begin{figure}
  \centering
  \includegraphics[width=0.48\textwidth]{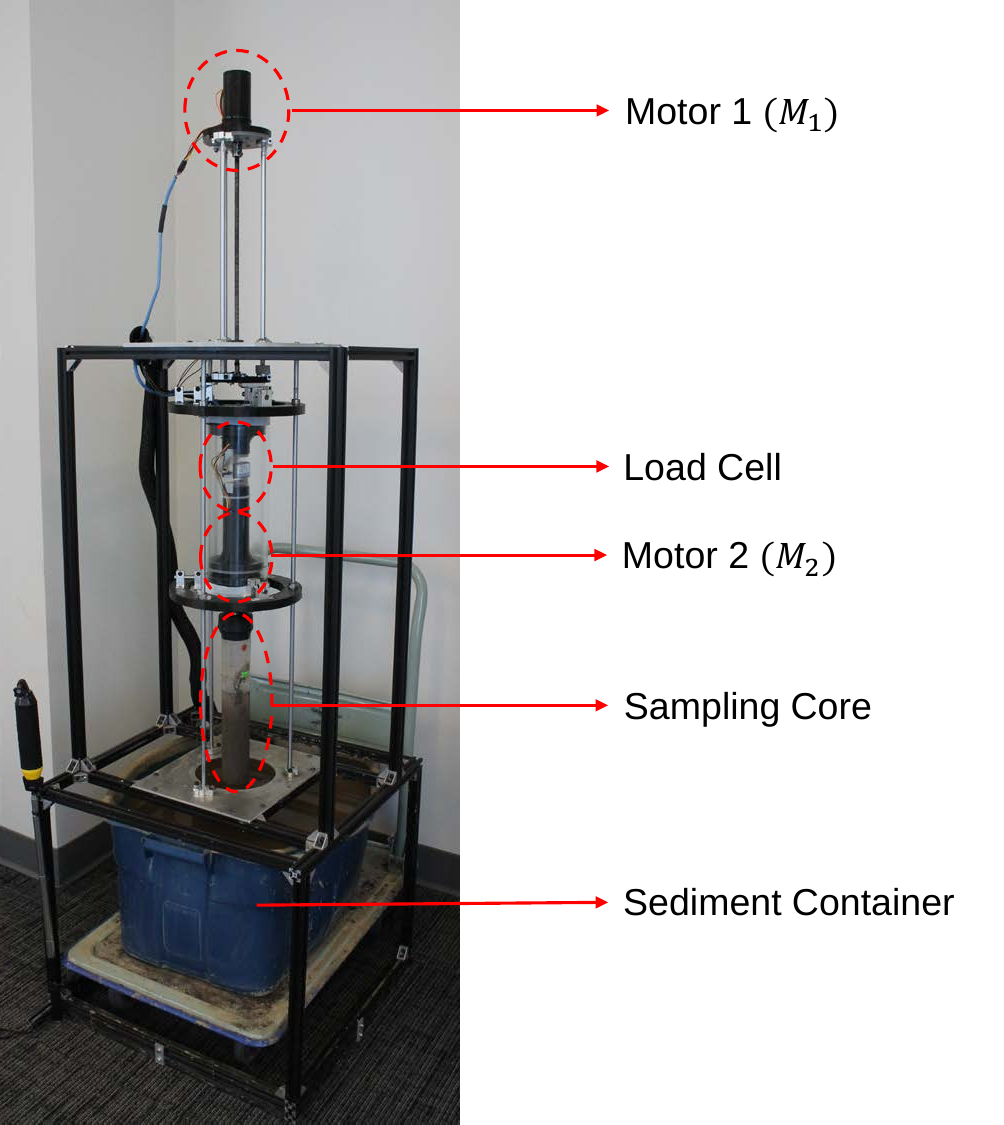}  
  \caption{A sediment sampling platform composed of two 12 V DC motors (Motor 1 ($M_{1}$) and Motor 2 ($M_{2}$)) with encoder, load cell, sampling core, and sediment container. The motion of the sampling core is based on the combination of $M_{1}$ and $M_{2}$; $M_{1}$ generates a linear motion, and $M_{2}$ generates rotary motion. The load cell measures the force during the sampling process.}
  \label{test-bed} 
\end{figure}

Our goal was to evaluate and optimize sediment sampling methods for a robotic sediment sampling system that is capable of collecting minimally-disturbed sediment cores for a range of sediment types. We developed and tested a robotic sampling platform in the laboratory to test functionality under a range of sediment types and operating conditions. Specifically, we focused on three patterns by which the coring device was driven into the sediment (linear, helical, and an oscillating zig-zag motion) for three sediment types (coarse sand, medium sand, and silt). The general configuration of the sediment sampling platform, including its two 12 V DC motors, load cell, sampling core, and sediment container is shown in Fig. \ref{test-bed}.

\begin{figure*}
\centering
  \begin{subfigure}[b]{0.30\linewidth}
    \centering
    \includegraphics[width=0.9\linewidth]{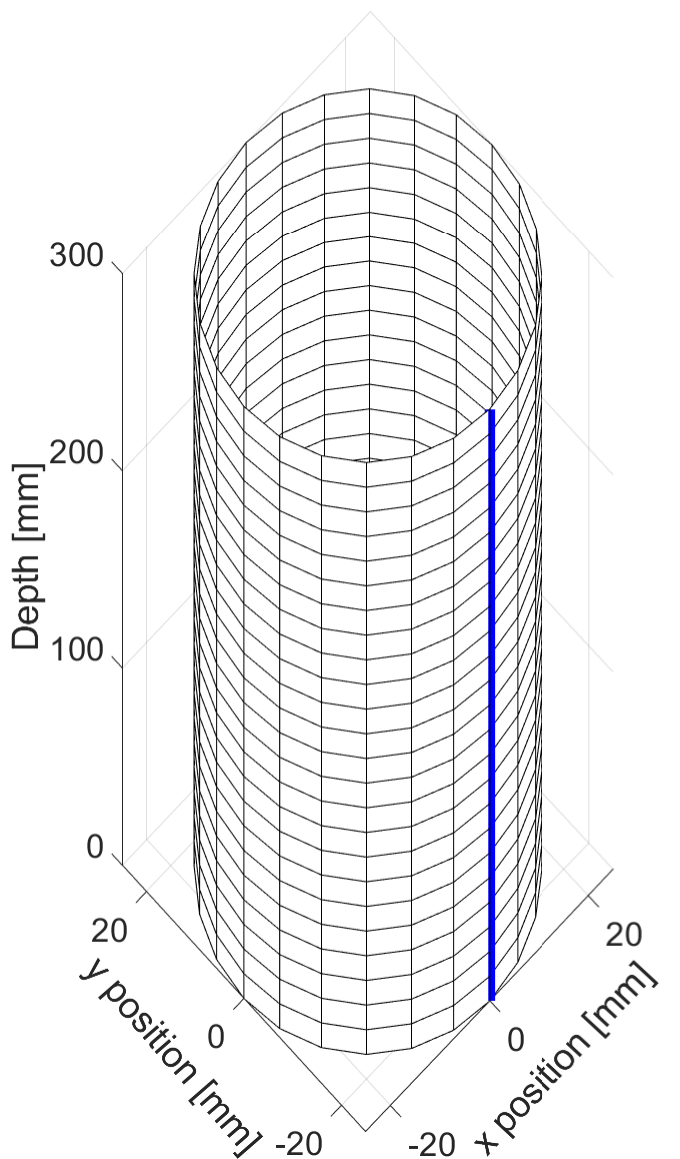} 
    \caption{Linear} 
    \label{patterns:Linear Motion} 
  \end{subfigure}
  \begin{subfigure}[b]{0.30\linewidth}
    \centering
    \includegraphics[width=0.9\linewidth]{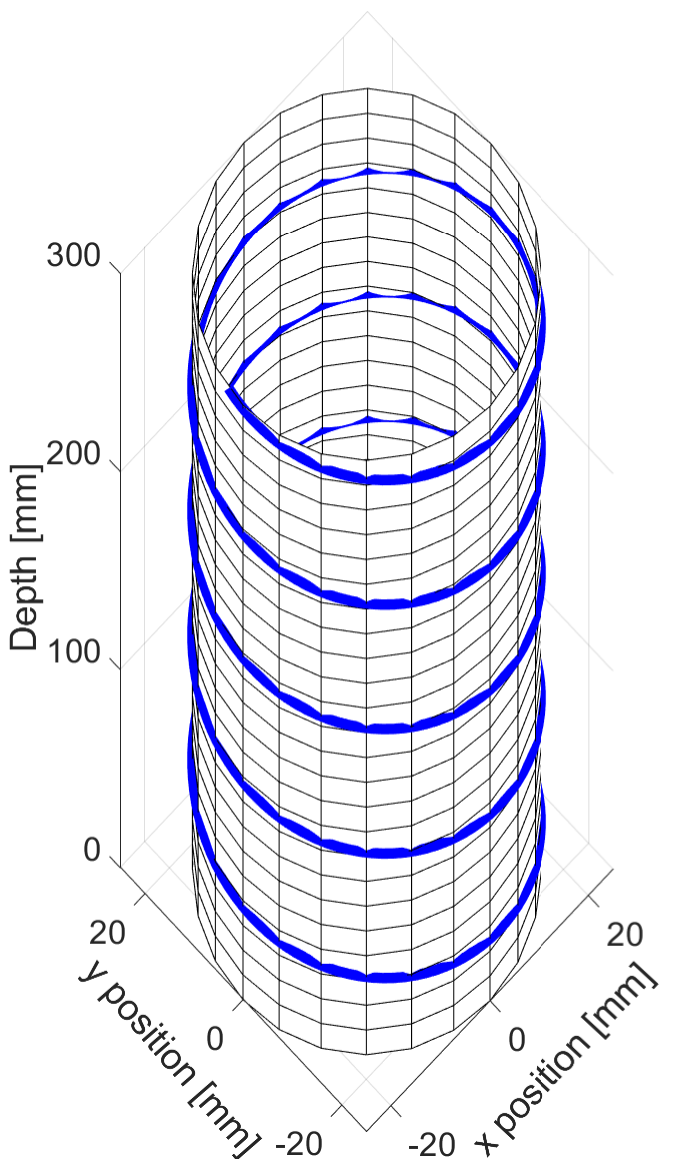} 
    \caption{Helical} 
    \label{patterns:Helical Motion} 
  \end{subfigure} 
  \begin{subfigure}[b]{0.30\linewidth}
    \centering
    \includegraphics[width=0.9\linewidth]{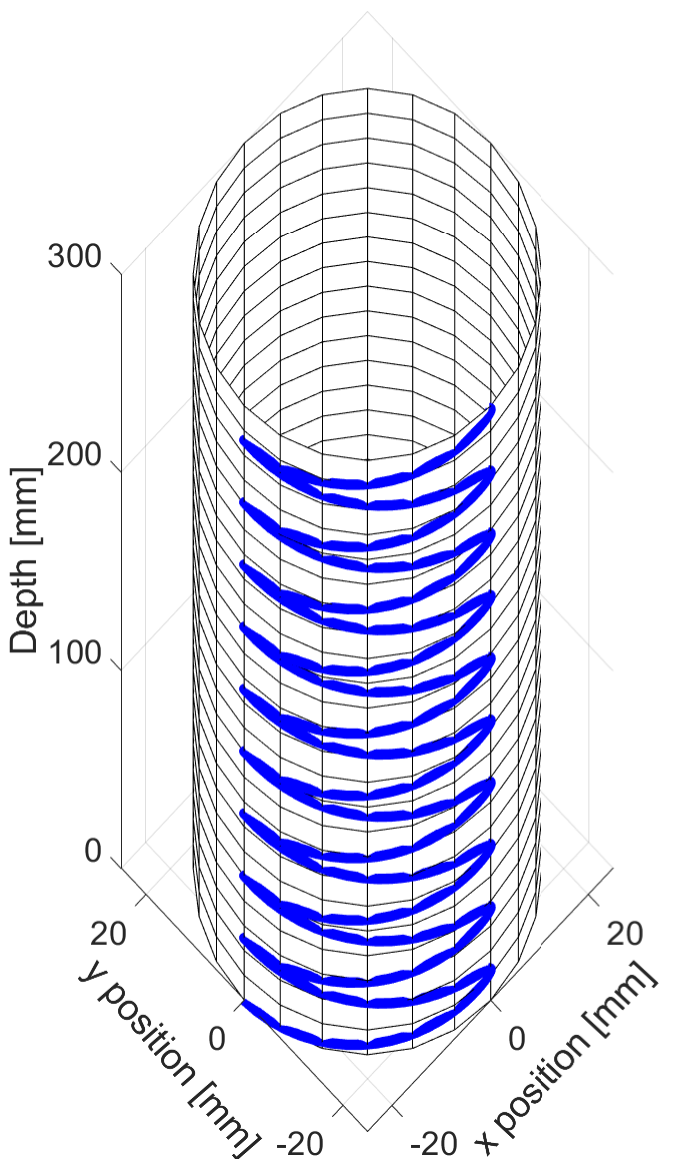} 
    \caption{Zig-zag} 
    \label{patterns:Zig-Zag Motion} 
  \end{subfigure}
\caption{The sediment sampling patterns applied in this study: (a) Linear motion, (b) Helical motion, and (c) Zig-zag motion.}
\label{patterns} 
\end{figure*}

\subsection{Sediment Sampling Platform Specifications}
The sampling patterns of the sediment sampling platform are based on a combination of motions by Motor 1 ($M_{1}$) generating a linear motion and Motor 2 ($M_{2}$) generating a rotary motion. The $M_{1}$ encoder measures the linear velocity of the sediment sampling platform while the $M_{2}$ encoder measures the angular velocity of the sampling core. A commercial plastic polyvinyl chloride (PVC) core liner for sediment sampling with an outer diameter of 50.8 mm (inner diameter of 47.8 mm  and wall thickness of 1.5 mm) and a length of 304.8 mm was used as the sampling core liner. A one-way check valve was installed on the core liner to prevent the sampled sediment from flushing out of the core liner upon retrieval. A sediment container is located on the bottom of the sediment sampling platform. The signals monitored by the sediment sampling platform are shown in Table \ref{Table1a} and the specifications of the sediment sampling platform are shown in Table \ref{Table1b}. The load cell measures the penetration force during the sampling process. Current sensors are installed to measure the input current of each motor. Specifically, this platform has a maximum (no-load) linear velocity of 38 mm/s and angular velocity of 12  rad/s. The maximum penetration force is 15 kg and the current limit is 3 A. Its vertical distance range is 0 to 300 mm.

\begin {table}
\centering
\caption {Signals monitored by the sediment sampling platform and specifications of the sediment sampling platform.}
\label{Table1}
  \begin{subtable}{1\linewidth}
    \begin{center}  
  \caption{Monitoring signals}
  \vspace{-1ex}
  \label{Table1a}
    \begin{tabular}{|c|c|}
      \hline
      Measuring sensor & Signal monitored  \\ \hline
      \hline
      Motor 1 ($M_{1}$) encoder & Linear velocity $v$  \\ \hline
      Motor 2 ($M_{2}$) encoder & Angular velocity $\omega_{r}$   \\ \hline
      Load cell & Penetration force $F_{p}$\\ \hline    
      Current sensor & $M_{1}$ \& $M_{2}$ current $C_1, C_2$\\ \hline
    \end{tabular}
  \end{center}  
  \end{subtable}

\vspace{3ex}
  \begin {subtable}{1\linewidth}
  \begin{center}  
    \caption{Specification}
    \vspace{-1ex}
  \label{Table1b} 
    \begin{tabular}{|c|c|}
      \hline
      Specification [unit] & Range \\ \hline
      \hline
      Linear $M_{1}$ velocity [mm/s] & 0 - 38\\ \hline 
      Rotational $M_{2}$ speed [rad/s] & 0 - 12  \\ \hline
      Rotational $M_{2}$ frequency [Hz] & 0 - 50   \\ \hline  
      Penetration force [kg] & 0 - 15   \\ \hline    
      Driving distance [mm] & 0 - 300  \\ \hline
      Current sensor [A] & 0 - 3\\ \hline
    \end{tabular}
  \end{center}
  \end{subtable}
\end {table}

\subsection{Sampling Pattern}
We applied three different sample sediment coring patterns based on our empirical studies: linear, helical, and an oscillating zig-zag motion as depicted in Fig. \ref{patterns}. The linear motion is the core samplers’ default pattern. The helical motion is the drilling pattern made by manual ground-drilling augers. The zig-zag motion rotates the core liner left and right, recursively. A demonstration of these patterns can be found from our experiment video at \url{https://youtu.be/W8gBe9SDXNw}.  

Although only three parameterized patterns are explored in this study, the sediment sampling platform can generate an arbitrary blend of the two motions induced by motors $M_{1}$ and $M_{2}$; $M_{2}$ is connected directly to the sampling core for rotary motion, and $M_{1}$ drives the combined corer plus $M_{2}$ mechanism up-and-down for linear motion. The linear motion (Fig. \ref{patterns:Linear Motion}) generated by $M_{1}$ is vertical motion without rotation, i.e., $\omega _{r}=0$. The helical motion (Fig. \ref{patterns:Helical Motion}) and zig-zag motion (Fig. \ref{patterns:Zig-Zag Motion}) are the combinations of $M_{1}$ and $M_{2}$. The helical motion is generated by combining linear motion while the sampling core is rotating. The zig-zag motion is composed of two different helical motions: both right-hand and left-hand motions combined with the linear motion to drive the coring tube into the sediment. The changing rate of the rotational motion direction depends on the angular velocity $\omega _{r}$ and the frequency of the motor input signal $f_{r}$. 

\begin{figure*}[tbp] 
\centering
  \begin{subfigure}[b]{0.9\linewidth}
    \centering
    \includegraphics[width=1\linewidth]{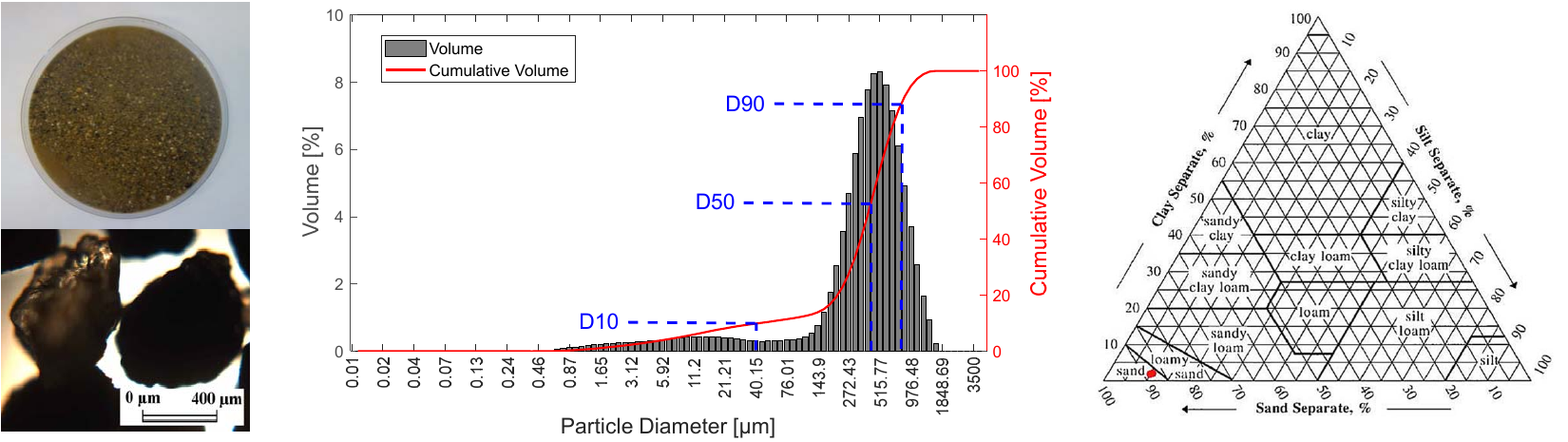} 
    \caption{Coarse sand}  
  \end{subfigure} 
  \begin{subfigure}[b]{0.9\linewidth}
    \centering
    \includegraphics[width=1\linewidth]{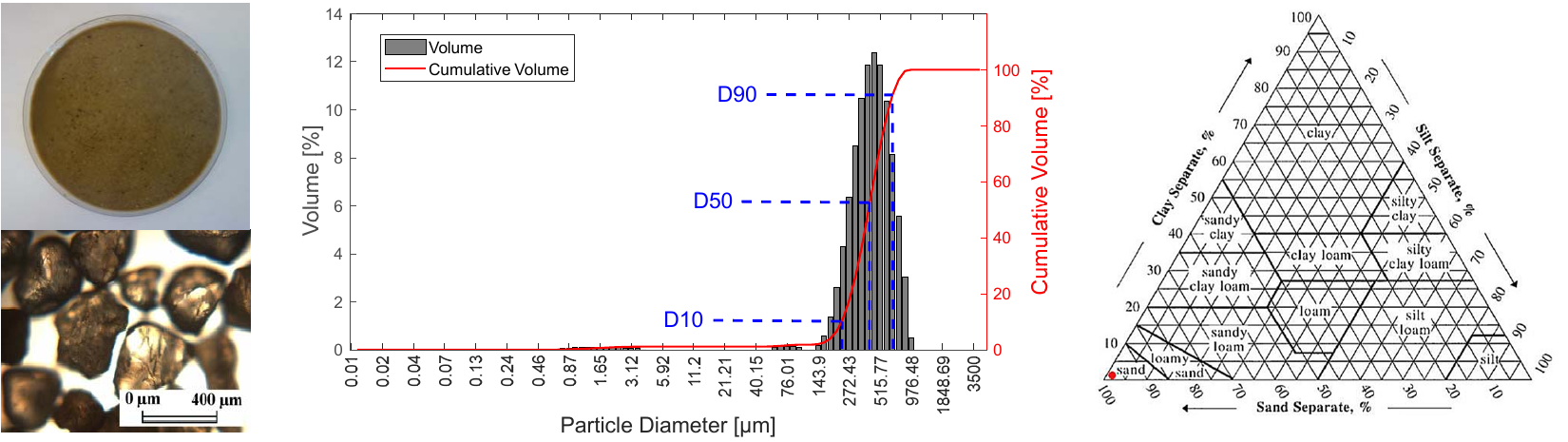} 
    \caption{Medium sand} 
  \end{subfigure} 
  \begin{subfigure}[b]{0.9\linewidth}
    \centering
    \includegraphics[width=1\linewidth]{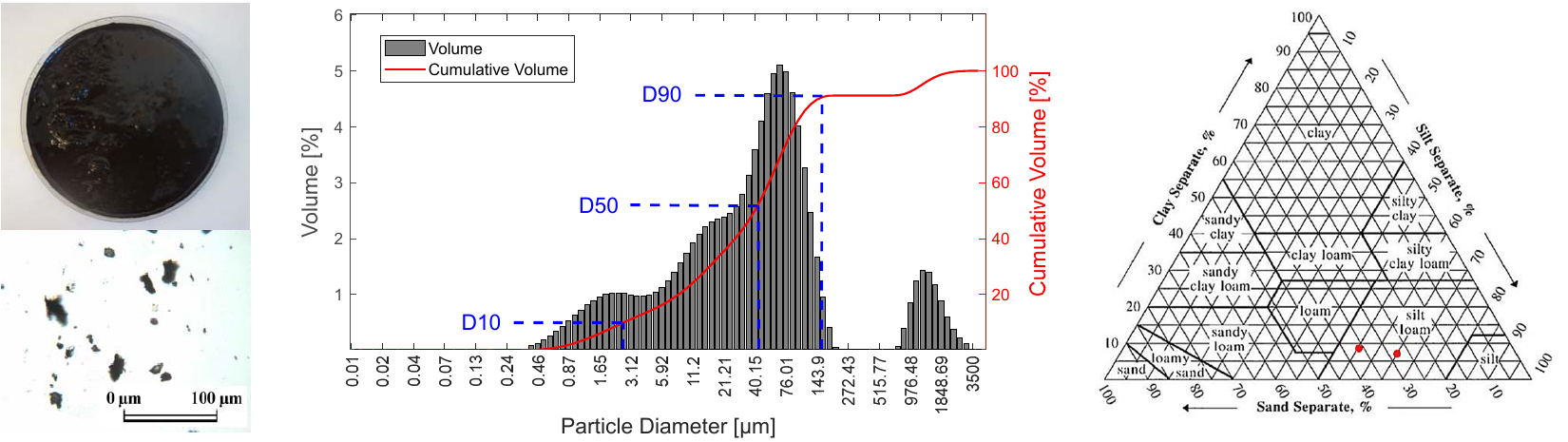} 
    \caption{Silt}  
  \end{subfigure}
  
\caption{Types of sediments: (a) Coarse sand (D50 $=$ \SI{409.85}{\micro\meter}), (b) Medium sand (D50 $=$ \SI{408.58}{\micro\meter}), and (c) Silt (D50 $=$ \SI{45.26}{\micro\meter}). The top panel shows each sediment type at $100 \%$ saturation in a petri dish, and the bottom panel is a microscopic view of particles observed using a polarizing microscope (\textit{Leitz Laborlux 12 POL S}) (left). The particle distribution of each sediment (middle). Sediment classification ternary diagrams depict the texture of each sediment based on the particle-size distribution results and red dots indicate the type of the sediment using the USDA soil texture calculator (right).}
\label{sediment texture} 
\end{figure*}

Let $p(x, y, z)$ be the arbitrary point on the surface of the sampling core. The position of the point can be then expressed as Eq. \ref{Equation1}, \ref{Equation2}, and \ref{Equation3}, respectively, depending on a motion where $r$ is the radius of the sampling core, $\omega _{r}$ is the angular velocity of the sampling core, and $v$ is the feed rate, which is the linear velocity of the sampling core:\\   

\textit{Pattern 1 ($P_{1}$): Linear motion}
\begin{equation}
\label{Equation1}
\left\{\begin{matrix}
x=rsin(\omega _{r}t)=0\\ 
y=rcos(\omega _{r}t)=r\\
z=-vt 
\end{matrix}\right.
\end{equation}

\textit{Pattern 2 ($P_{2}$): Helical motion}
\begin{equation}
\label{Equation2}
\left\{\begin{matrix}
x=rsin(\omega _{r}t)\\ 
y=rcos(\omega _{r}t)\\
z=-vt 
\end{matrix}\right.
\end{equation}

\textit{Pattern 3 ($P_{3}$): Zig-zag motion}
\begin{equation}
\label{Equation3}
\left\{\begin{matrix}
x=rsin({10\omega _{r}\left | sin(0.1f_{r}t)) \right |}/{f_{r}})\\ 
y=rcos({10\omega _{r}\left | sin(0.1ft)) \right |}/{f_{r}})\\
z=-vt 
\end{matrix}\right.
\end{equation}

\subsection{Classification of Sediment}

We used three types of sediment classified by the particle-size distribution \cite{blott_gradistat:_2001}\cite{blott_particle_2012} as shown in Fig. \ref{sediment texture}. We defined sample sediments (i.e., coarse sand, medium sand, and silt) based on the United States Department of Agriculture (USDA) soil texture classes and subclasses, and used a laser particle counter (Mastersizer 3000, Malvern Panalytical, Malvern, UK) to provide precise particle distribution of the sediments.

\section{Multiple Objective Optimization}
\label{sec:methodology}
We defined the optimal sediment sampling pattern as the one that collects the greatest amount of sediment with greater work efficiency. Therefore, we consider this as a multiple objective optimization problem in order to find the optimal sediment sampling pattern based on the type of sediment. A common approach to optimize the multi-objective problem is to minimize the sum of the individual objectives using the weighted-sum method \cite{ngatchou_pareto_2005}\cite{vanderplaats_multi-objective_nodate}. A weighting factor indicates the importance of each objective. We implement this as a discrete optimization problem because we are limited in the number of design iterations of the physical systems that we are able to create and test. While we recognize there are an infinite set of potential solutions, it is impossible to run the experiment iteratively by continuously increasing the input parameters. Hence, we can express the objective function as:
\begin{subequations}
\begin{align}
\text{Minimize:~} & F(\mathbb{X})=w_{1}f_{1}(\mathbb{X})+w_{2}f_{2}(\mathbb{X}) \\
    \text{Subject to:~} & x_{i}\in \mathbb{X}, (i=1,2,3) \\
     & w_{1,2} \geq 0 \text{~and~} w_{1}+w_{2}=1  \\
     & C_{1,2} \leq  C_{max}
\end{align}
\label{Eq5}
\end{subequations}
\noindent where $\mathbb{X}$ is a finite set,
\begin{equation}
\mathbb{X}=
\begin{bmatrix}
x_{1}\\ 
x_{2}\\ 
x_{3}
\end{bmatrix}=
\begin{bmatrix}
v\\ 
\omega_{r}\\ 
f_{r}
\end{bmatrix}
\label{EqX}
\end{equation}

\noindent where  $x_{1}=v$ is the linear velocity [mm/s] of $M_1$, $x_{2}=\omega_{r}$ is the angular velocity [rad/s] of $M_2$, and $x_{3}=f_{r}$ is the direction changing frequency [Hz] of $M_2$. In Eq. \ref{Eq5}, $w_{1}$ and $w_{2}$ are finite weighting factors; $C_{1}$ and $C_{2}$ are the values of the electrical current consumed by $M_1$ and $M_2$, respectively; and $C_{max}$ is the value of the maximum current allowed for each motor that prevents the motors and systems from overload damage. We can express $f_{1}(\mathbb{X})$ and $f_{2}(\mathbb{X})$ in the following functions:
\begin{equation}
\label{Equation4}
f_{1}(\mathbb{X})= \left (\frac{m_{s}(x_{1}, x_{2}, x_{3})}{V_{d}} \right)^{-1}
\end{equation}
\begin{equation}
\label{Equation5}
f_{2}(\mathbb{X})= \left (\frac{\int F_{p}(x_{1}, x_{2}, x_{3}) ds}{t} \right)^{-1}
\end{equation}

\noindent where $f_{1}(\mathbb{X})$ is a reciprocal function of the density of the sampled sediment, and $f_{2}(\mathbb{X})$ is a reciprocal function of the power of the sediment sampler. The goal of the experiment is to minimize $f_{1}(\mathbb{X})$ and $f_{2}(\mathbb{X})$. Minimizing $f_{1}(\mathbb{X})$ and $f_{2}(\mathbb{X})$ identifies the pattern with the largest sampled mass and the most efficient work performed by the sampler. The variable $m_{s}$ is the mass of sampled sediment, and ${V_{d}}=\pi r^{2}{L_{d}}$ is the desired volume of the sampling core based on the core radius $r$ and the desired depth ${L_{d}}$. $F_{p}$ is the penetration force of the sampling core based on the load cell data, and $s$ is the total distance traveled by the sampling core from the $M_2$ encoder. Based on the sets of $f_{1}(\mathbb{X})$ and $f_{2}(\mathbb{X})$ values, we can apply multiple objective optimization to identify the optimal pattern for a given sediment type, sample mass, etc. The optimal pattern is somewhat dependent on the chosen weights $w_{1}$ and $w_{2}$ that reflect the user's emphasis on sample mass versus efficiency.

\section{Experiment}
\label{sec:experiment}
For the experimental study, we used three sampling patterns and three sediment types to find the optimal sampling pattern for each type of sediment. We conducted independent experiments that varied the linear and rotational velocities for linear and helical sampling patterns and the frequency of the back and forth motion (direction of the rotation) for the zig-zag sampling pattern. We set $C_{max}$~$= 3$ A as a value of the maximum current allowed to each motor. In addition, we applied the same compaction protocol to each sediment in every iteration to conduct experiment under the same condition.

\subsection{Experiment Procedure}
\begin{figure*}[t!] 
\centering
  \begin{subfigure}[b]{\linewidth}
    \centering
    \includegraphics[width=0.6\linewidth]{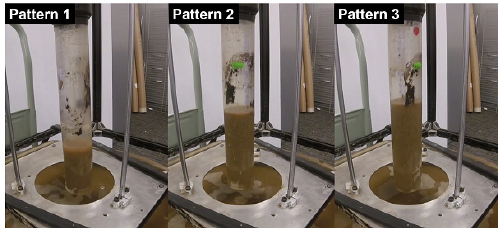} 
    \caption{Coarse sand} 
    \label{mass_p1} 
  \end{subfigure} 
  \begin{subfigure}[b]{\linewidth}
    \centering
 \includegraphics[width=0.6\linewidth]{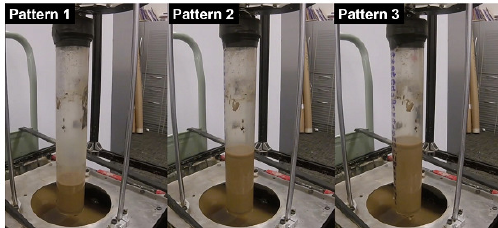} 
    \caption{Medium sand} 
    \label{mass_p2} 
  \end{subfigure} 
  \begin{subfigure}[b]{\linewidth}
    \centering
    \includegraphics[width=0.6\linewidth]{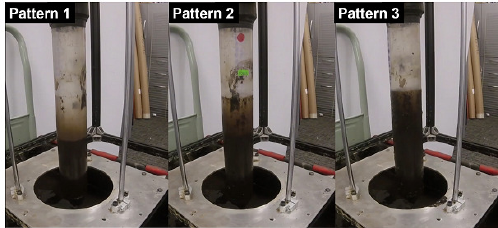} 
    \caption{Silt} 
    \label{mass_p3} 
  \end{subfigure}
\caption{Recovered sediment for each sediment sampling coring approach (i.e., pattern) and sediment type. Pattern 1 used a linear motion, Pattern 2 used a helical motion, and Pattern 3 used an oscillating zig-zag motion. Video recordings of the experiments are available at \url{https://youtu.be/W8gBe9SDXNw}.}
\label{sample example} 
\end{figure*}

We utilized the two-step experiment: Step 1 -- Apply two patterns $P_{1}$ (linear) and $P_{2}$ (helix); and Step 2 -- Apply $P_{3}$ (zig-zag) based on the selected (statistically significant) patterns from the Step 1. We applied this procedure to all three sediment types. The recovered sediment from each experiment is shown in Fig. \ref{sample example}.

For Step 1, we applied 20 combination sets ($4$ patterns for $P_{1}$ and $16$ patterns for $P_{2}$) of $M_1$ and $M_2$ inputs: $v \in \left \{ 15, 22, 29, 38\right \}$ and $\omega_{r} \in \left \{ 0, 3, 6, 9, 12\right \}$. The sequence for each experimental combination was based on the Simple Random Sampling (SRS) \cite{moore1996basic} method to avoid bias. We ran three trials for each combination. For example, in the case of $P_{2}$, we could have 16 combinations, and for each combination, we repeated it three times. As a result, we had 48 sampling results of $P_{2}$ for each sediment. We collected four pieces of data: penetration depth, sampled sediment mass, penetration force, and motor current. We weighed sediment samples manually using a top load balance and other data were based on sensors. 

For Step 2, we applied $P_{3}$ to validate the zig-zag motion. As shown in Eq. \ref{Equation3}, the $P_{3}$ sampling core changes its rotating direction depending on the frequency of the motor input signal $f_{r}$, which is the direction change rate. Because we used the optimal angular velocity $\omega_{r}$ from our Step 1 results, we changed only the input frequency, $f_{r} \in \left \{ 0, 10, 30, 50\right \}$. 

\subsection{Statistical Approach}
Based on Step 1 experimental results, $f_{1}(\mathbb{X})$ and $f_{2}(\mathbb{X})$ can be calculated as shown in Table \ref{Table4}, where $f_{1}(\mathbb{X})$ and $f_{2}(\mathbb{X})$ are normalized values into the range $[1, 10]$. First, we ran a two-way ANOVA to verify the significance of the results from two motors. Second, in order to select the patterns to use in the Step 2 experiment, we first found the patterns representing the minimum values in $P_{1}$ and $P_{2}$ (see bolded values in Table \ref{Table4}). We then used a multiple comparison method \cite{KutnerMichaelH.2005Alsm} to find statistically significantly different patterns of these patterns, and we chose the results as the final patterns for the Step 2 experiment.

Step 2 experimental results based on the selected values from Step 1 experiment results are shown in Table \ref{Table5}, where $f_{1}(\mathbb{X})$ and $f_{2}(\mathbb{X})$ values are normalized values into the range $[1, 10]$. We also ran a two-way ANOVA to verify the significance of the selected patterns and the direction change rate. Based on Step 1 and Step 2 experimental results, we applied a weighted-sum multiple objective optimization to find the optimal pattern depending on the sediment type and weight configuration.

\subsection{Experiment Result: Step 1}
\label{step1}
As we described in Section \ref{sec:methodology}, $f_{1}(\mathbb{X})$ is a reciprocal function of the density of the sampled sediment based on the desired volume. $f_{2}(\mathbb{X})$ is a reciprocal function of the power of the sediment sampler, which indicates the sampling efficiency (the penetration force times the travel time to the desired depth). The lowest $f_{1}(\mathbb{X})$ indicates the largest amount of sediment sampled and the lowest $f_{2}(\mathbb{X})$ indicates the pattern with the highest work efficiency.

\begin{table*}
\centering
\caption {Experiment results of Step 1, $P_{1}$ and $P_{2}$: (a) Coarse sand, (b) Medium sand, and (c) Silt. The minimum values for each coring pattern and sediment type are highlighted in bold. The minimum $f_{1}(\mathbb{X})$ indicates the largest sampled mass, and the minimum $f_{2}(\mathbb{X})$ is the best work efficiency.}
\label{Table4}

\begin{subtable}{\linewidth}
\begin{center}
\caption {Coarse Sand}
  \vspace{-1ex}
\label{Table4a}
\begin{adjustbox}{width=0.8\textwidth}
\begin{tabular}{|c|c|c|c|c|c|c|c|c|c|c|c|} \hline
\centering
Objective & \multicolumn{5}{c|}{$f_{1}(\mathbb{X})$} & \multicolumn{5}{c|}{$f_{2}(\mathbb{X})$} \\ \hline
Pattern & $P_{1}$ & \multicolumn{4}{c|}{$P_{2}$} & $P_{1}$ & \multicolumn{4}{c|}{$P_{2}$} \\ 
\hline \hline
\backslashbox{$v$[mm/s]}{$\omega_{r}$[rad/s]} & 0 & 3 & 6 & 9 & 12 & 0 & 3 & 6 & 9 & 12\\ \hline 
15 & 0.00 & 0.00 & 4.69 & 4.18 & 3.10 & 4.55 & 8.42 & 7.02 & 5.75 & 4.80 	\\ \hline
22 & 4.93 & 0.00 & 6.15 & 4.03 & 3.34 & 2.36 & 3.54 & 4.63 & 3.48 & 2.73 	\\ \hline
29 & 6.97 & 7.30 & 6.97 & 2.96 & 3.17 & 1.51 & 1.79 & 4.83 & 2.09 & 2.30 	\\ \hline
38 & \textbf{4.67} & 4.66 & 10.00 & 6.50 & \textbf{2.63} & \textbf{1.11} & 1.40 & \textbf{1.35} & 1.78 & 1.37 	\\ \hline
\end{tabular}
\end{adjustbox}
\end{center}
\end{subtable} 

\vspace{3ex}
\begin{subtable}{\linewidth}
\begin{center}
\caption {Medium Sand}
  \vspace{-1ex}
\label{Table4b}
\begin{adjustbox}{width=0.8\textwidth}
\begin{tabular}{|c|c|c|c|c|c|c|c|c|c|c|c|} \hline
\centering
Objective & \multicolumn{5}{c|}{$f_{1}(\mathbb{X})$} & \multicolumn{5}{c|}{$f_{2}(\mathbb{X})$} \\ \hline
Pattern & $P_{1}$ & \multicolumn{4}{c|}{$P_{2}$} & $P_{1}$ & \multicolumn{4}{c|}{$P_{2}$} \\ \hline \hline
\backslashbox{$v$[mm/s]}{$\omega_{r}$[rad/s]} & 0 & 3 & 6 & 9 & 12 & 0 & 3 & 6 & 9 & 12\\ \hline 
15 & 10.00 & 9.96 & 8.12 & 4.12 & 4.87 & 7.39 & 10.00 & 8.01 & 4.69 & 6.21 	\\ \hline
22 & 7.59 & 9.28 & 8.06 & 5.41 & 3.99 & 4.05 & 3.29 & 5.46 & 5.11 & 2.30 	\\ \hline
29 & 6.45 & 8.46 & 6.56 & 4.80 & \textbf{1.41} & 1.96 & 2.16 & 2.65 & 1.71 & \textbf{1.00} 	\\ \hline
38 & \textbf{6.30} & 8.46 & 7.11 & 6.56 & 5.27 & \textbf{1.12} & 1.56 & 2.00 & 1.21 & 1.99 	\\ \hline
\end{tabular}
\end{adjustbox}
\end{center}
\end{subtable} 

\vspace{3ex}
\begin{subtable}{\linewidth}
\begin{center}
  \caption {Silt}
\vspace{-1ex}  
\label{Table4c}
\begin{adjustbox}{width=0.8\textwidth}
\begin{tabular}{|c|c|c|c|c|c|c|c|c|c|c|c|} \hline
\centering
Objective & \multicolumn{5}{c|}{$f_{1}(\mathbb{X})$} & \multicolumn{5}{c|}{$f_{2}(\mathbb{X})$} \\ \hline
Pattern & $P_{1}$ & \multicolumn{4}{c|}{$P_{2}$} & $P_{1}$ & \multicolumn{4}{c|}{$P_{2}$} \\ \hline \hline
\backslashbox{$v$[mm/s]}{$\omega_{r}$[rad/s]} & 0 & 3 & 6 & 9 & 12 & 0 & 3 & 6 & 9 & 12\\ \hline 
15 & 7.51 & 10.00 & 3.07 & 2.82 & \textbf{2.46} & 5.61 & 4.90 & 7.08 & 8.21 & 9.39 	\\ \hline
22 & 4.66 & 3.80 & 3.07 & 2.58 & 2.95 & 2.22 & 3.85 & 4.70 & 4.62 & 3.11 	\\ \hline
29 & \textbf{4.41} & 4.41 & 3.55 & 3.43 & 2.82 & \textbf{1.00} & 2.70 & 2.48 & 3.49 & 2.25 	\\ \hline
38 & 5.20 & 5.62 & 4.04 & 4.04 & 3.92 & 1.08 & \textbf{1.94} & 2.25 & 2.53 & 2.54 	\\ \hline
\end{tabular}
\end{adjustbox}
\end{center}
\end{subtable}
\end {table*}

In the case of coarse sand, not all null hypotheses were rejected, based on results (Table \ref{Table4a}). The velocity of $M_1$ ($v$[mm/s]) and the interaction between the velocities of $M_1$ and $M_2$ were significant ($p$-value $<$ 0.05). The velocity of $M_2$ ($\omega_{r}$[rad/s]) was not significant. To select the patterns of Step 2, the pattern with 38 mm/s of linear velocity and 12 rad/s of angular velocity ($P_{2}$:~$38/12/0$)\footnote[2]{We use notation $(P_{x}$:~$v/\omega_{r}/f_{r})$, where $x$ is the pattern number, $v$ is the linear velocity, $\omega_{r}$ is the angular velocity, and $f_{r}$ is the frequency.} provided the minimum value of $f_{1}(\mathbb{X})$. The mass of sediment (257.67 g) with this pattern was the largest amount recovered out of the 20 patterns. A value of 0 means no sample was collected, so we excluded these data from the optimization process. 

In the case of medium sand, all null hypotheses were rejected based on the results (Table \ref{Table4b}). The velocities of $M_1$ and $M_2$, as well as the interaction between the velocities of $M_1$ and $M_2$, were significant ($p$-value $<$ 0.05). The pattern with 29 mm/s of linear velocity and 12 rad/s of angular velocity ($P_{2}$:~$29/12/0$) provided the minimum $f_{1}(\mathbb{X})$ value. The mass of sediment (458.33 g) from this pattern was the largest amount out of the 20 patterns. 

In the case of silt, all null hypotheses were rejected based on the results (Table \ref{Table4c}). The velocities of $M_1$, $M_2$ and the interaction between the velocities of $M_1$, $M_2$ were significant ($p$-value $<$ 0.05). The pattern with 15 mm/s of linear velocity and 12 rad/s of angular velocity ($P_{2}$:~$15/12/0$) provided the minimum value of $f_{1}(\mathbb{X})$. The mass of sediment (281 g) with this pattern was the largest amount out of the 20 patterns. 

\subsection{Experiment Result: Step 2}
\label{step2}
The patterns were selected based on the multiple comparison method to test $P_{3}$ for each sediment. Multiple patterns are selected for coarse sand (7 patterns) and silt (14 patterns). In the case of medium sand, only one pattern was selected because results of the pattern with 29 mm/s of linear velocity and 12 rad/s of angular velocity ($P_{2}$:~$29/12/0$) were significantly different from the other patterns. Similar to Table \ref{Table4}, the lowest values for each pattern and sediment are in bold.

In the case of coarse sand (Table \ref{Table5a}), all null hypotheses were rejected. The velocities of $M_1$, $M_2$, and the interaction between the velocities of $M_1$ and $M_2$ were significant ($p$-value $<$ 0.05). The pattern with 38 mm/s of linear velocity, 12 rad/s of angular velocity, and 30 Hz of the motor input frequency ($P_{3}$:~$38/12/30$) provided the minimum value of $f_{1}(\mathbb{X})$. The mass of sediment (486 g) with this pattern was the largest amount out of the 28 patterns. 

\begin{table*}
\centering
\caption {Experiment results of Step 2, $P_{3}$: (a) Coarse sand, (b) Medium sand, and (c) Silt. The minimum values are highlighted in bold. The minimum $f_{1}(\mathbb{X})$ indicates the largest sampled mass, and the minimum $f_{2}(\mathbb{X})$ is the best work efficiency.}
\label{Table5}

\begin{subtable}{\linewidth}
\begin{center}
\caption {Coarse sand}
  \vspace{-1ex}
\label{Table5a}
\begin{adjustbox}{width=0.65\textwidth}
\begin{tabular}{|c|c|c|c|c|c|c|c|c|c|} \hline
\centering
Objective & \multicolumn{4}{c|}{$f_{1}(\mathbb{X})$} & \multicolumn{4}{c|}{$f_{2}(\mathbb{X})$} \\ \hline
Pattern & $P_{2}$ & \multicolumn{3}{c|}{$P_{3}$} & $P_{2}$ & \multicolumn{3}{c|}{$P_{3}$} \\ \hline \hline
\backslashbox{$v/\omega_{r}$}{$f_{r}$[Hz]} & 0 & 10 & 30 & 50 & 0 & 10 & 30 & 50 \\ \hline 
15/9 & 4.18 & 2.10 & 2.08 & 2.97 & 5.75 & 3.45 & 3.15 & 5.26  	\\ \hline
15/12 & 3.10 & 2.46 & 2.46 & 2.67 & 4.80 & 4.29 & 4.44 & 10.00  	\\ \hline
22/9 & 4.03 & 2.17 & 2.38 & 3.01 & 3.48 & 2.12 & 1.98 & 1.85  	\\ \hline
22/12 & 3.34 & 2.35 & 2.17 & 2.69 & 2.73 & 2.37 & 1.95 & 2.22  	\\ \hline
29/9 & 2.96 & 2.17 & 1.98 & 2.46 & 2.09 & 1.57 & 1.72 & 1.85   \\ \hline
29/12 & 3.17 & 1.91 & 1.91 & 2.55 & 2.30 & 1.39 & 1.55 & 1.50   \\ \hline
38/12 & \textbf{2.63} & 2.07 & \textbf{1.87} & 2.84 & \textbf{1.37} & \textbf{1.00} & 1.05 & 1.38  	\\ \hline
\end{tabular}
\end{adjustbox}
\end{center}
\end{subtable} 

\vspace{3ex}
\begin{subtable}{\linewidth}
\begin{center}
  \caption {Medium Sand}
\vspace{-1ex}  
\label{Table5b}
\begin{adjustbox}{width=0.65\textwidth}
\begin{tabular}{|c|c|c|c|c|c|c|c|c|c|} \hline
\centering
Objective & \multicolumn{4}{c|}{$f_{1}(\mathbb{X})$} & \multicolumn{4}{c|}{$f_{2}(\mathbb{X})$} \\ \hline
Pattern & $P_{2}$ & \multicolumn{3}{c|}{$P_{3}$} & $P_{2}$ & \multicolumn{3}{c|}{$P_{3}$} \\ \hline \hline
\backslashbox{$v/\omega_{r}$}{$f_{r}$[Hz]} & 0 & 10 & 30 & 50 & 0 & 10 & 30 & 50 \\ \hline 
29/12 & \textbf{1.41} & 1.07 & \textbf{1.00} & \textbf{1.00} & \textbf{1.00} & \textbf{3.30} & 3.39 & 3.73  	\\ \hline
\end{tabular}
\end{adjustbox}
\end{center}
\end{subtable} 

\vspace{3ex}
\begin{subtable}{\linewidth}
\begin{center}
\caption {Silt}
\vspace{-1ex}
\label{Table5c}
\begin{adjustbox}{width=0.65\textwidth}
\begin{tabular}{|c|c|c|c|c|c|c|c|c|c|} \hline
\centering
Objective & \multicolumn{4}{c|}{$f_{1}(\mathbb{X})$} & \multicolumn{4}{c|}{$f_{2}(\mathbb{X})$} \\ \hline
Pattern & $P_{2}$ & \multicolumn{3}{c|}{$P_{3}$} & $P_{2}$ & \multicolumn{3}{c|}{$P_{3}$} \\ \hline \hline
\backslashbox{$v/\omega_{r}$}{$f_{r}$[Hz]} & 0 & 10 & 30 & 50 & 0 & 10 & 30 & 50 \\ \hline 
15/6 & 3.07 & \textbf{1.00} & 2.58 & 2.46 & 7.08 & 7.97 & 7.68 & 5.53  	\\ \hline
15/9 & 2.82 & 3.31 & 3.31 & 4.41 & 8.21 & 6.24 & 7.03 & 5.67  	\\ \hline
15/12 & \textbf{2.46} & 5.01 & 5.38 & 6.35 & 9.39 & 6.07 & 6.98 & 10.00  	\\ \hline
22/3 & 3.80 & 4.41 & 5.38 & 5.01 & 3.85 & 2.87 & 2.73 & 2.34  	\\ \hline
22/6 & 3.07 & 2.58 & 3.68 & 3.19 & 4.70 & 5.07 & 3.90 & 3.08  	\\ \hline
22/9 & 2.58 & 5.50 & 6.59 & 7.32 & 4.62 & 4.74 & 5.60 & 5.13  	\\ \hline
22/12 & 2.95 & 5.14 & 5.86 & 5.14 & 3.11 & 3.98 & 4.46 & 3.83  	\\ \hline
29/3 & 4.41 & 3.31 & 2.58 & 2.70 & 2.70 & 2.04 & 1.85 & 1.75   \\ \hline
29/6 & 3.55 & 4.04 & 4.28 & 5.38 & 2.48 & 2.78 & 2.52 & 2.13   \\ \hline
29/9 & 3.43 & 2.58 & 4.65 & 4.53 & 3.49 & 2.64 & 2.98 & 2.58   \\ \hline
29/12 & 2.82 & 1.73 & 2.70 & 3.19 & 2.25 & 3.57 & 3.94 & 2.99   \\ \hline
38/6 & 4.04 & 4.53 & 6.23 & 5.26 & \textbf{2.25} & 1.93 & 1.81 & \textbf{1.38}  	\\ \hline
38/9 & 4.04 & 3.92 & 4.89 & 5.01 & 2.53 & 1.79 & 1.95 & 1.70  	\\ \hline
38/12 & 3.92 & 5.14 & 6.11 & 5.38 & 2.54 & 2.06 & 1.89 & 1.72  	\\ \hline
\end{tabular}
\end{adjustbox}
\end{center}
\end{subtable} 
\end {table*}

In the case of medium sand (Table \ref{Table5b}), all null hypotheses were rejected. The velocities of $M_1$ and $M_2$ were significant ($p$-value $<$ 0.05). There was not much difference between the patterns in $f_{1}(\mathbb{X})$. The patterns with 29 mm/s of linear velocity, 12 rad/s of angular velocity, 30 Hz of the motor input frequency ($P_{3}$:~$29/12/30$) and 50 Hz of the motor input frequency ($P_{3}$:~$29/12/50$) provided the minimum value of $f_{1}(\mathbb{X})$. The masses of sediment (493.67 g and 491.33 g) were respectively sampled in two patterns. 

In the case of silt (Table \ref{Table5c}), all null hypotheses were rejected. The velocities of $M_1$, $M_2$ and the interaction between the velocities of $M_1$ and $M_2$ were significant ($p$-value $<$ 0.05). The pattern with 15 mm/s of linear velocity, 6 rad/s of angular velocity, and 10Hz of the motor input frequency ($P_{3}$:~$15/6/10$) provided the minimum value of $f_{1}(\mathbb{X})$. The mass of sediment (306.33 g) with this pattern was the largest amount out of the 56 patterns. 

\subsection{Summary}
To maximize the mass of the sampled sediment and the power efficiency of the sediment sampling platform, we minimized the values of $f_{1}(\mathbb{X})$ and $f_{2}(\mathbb{X})$, respectively. Three types of sediment were tested against three parameterized classes of core sampler motion under two different user objectives. While parameter variation was not exhaustive, we conclude from the data that it appears sufficient for the sediments explored and user objectives. 

\begin{figure}[t!] 
\centering
  \begin{subfigure}[b]{\linewidth}
    \centering
    \includegraphics[width=0.85\linewidth]{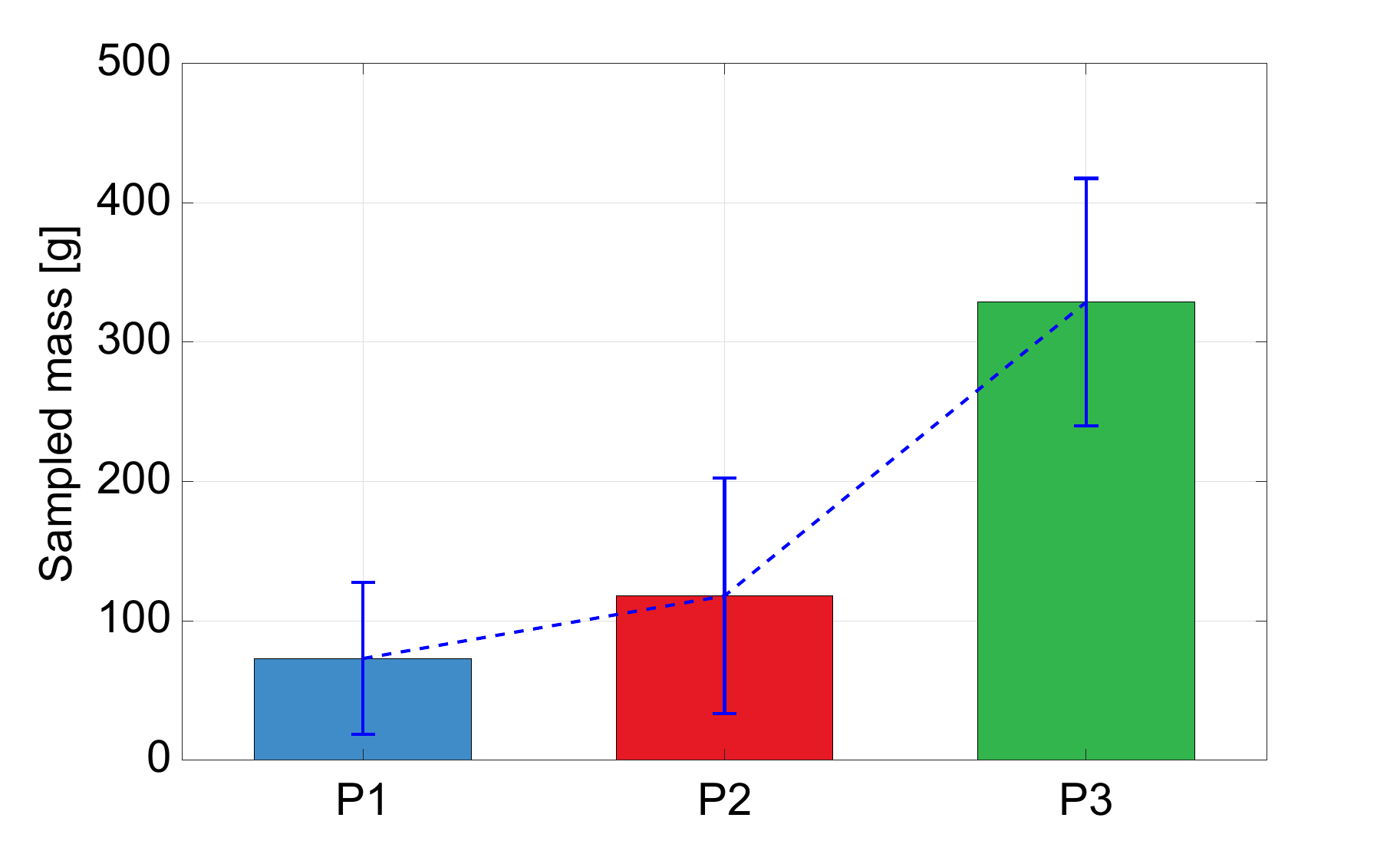}
\vspace{-1ex}    
    \caption{Coarse sand} 
    \label{mass_p1} 
  \end{subfigure} 
  \begin{subfigure}[b]{\linewidth}
    \centering
 	\includegraphics[width=0.85\linewidth]{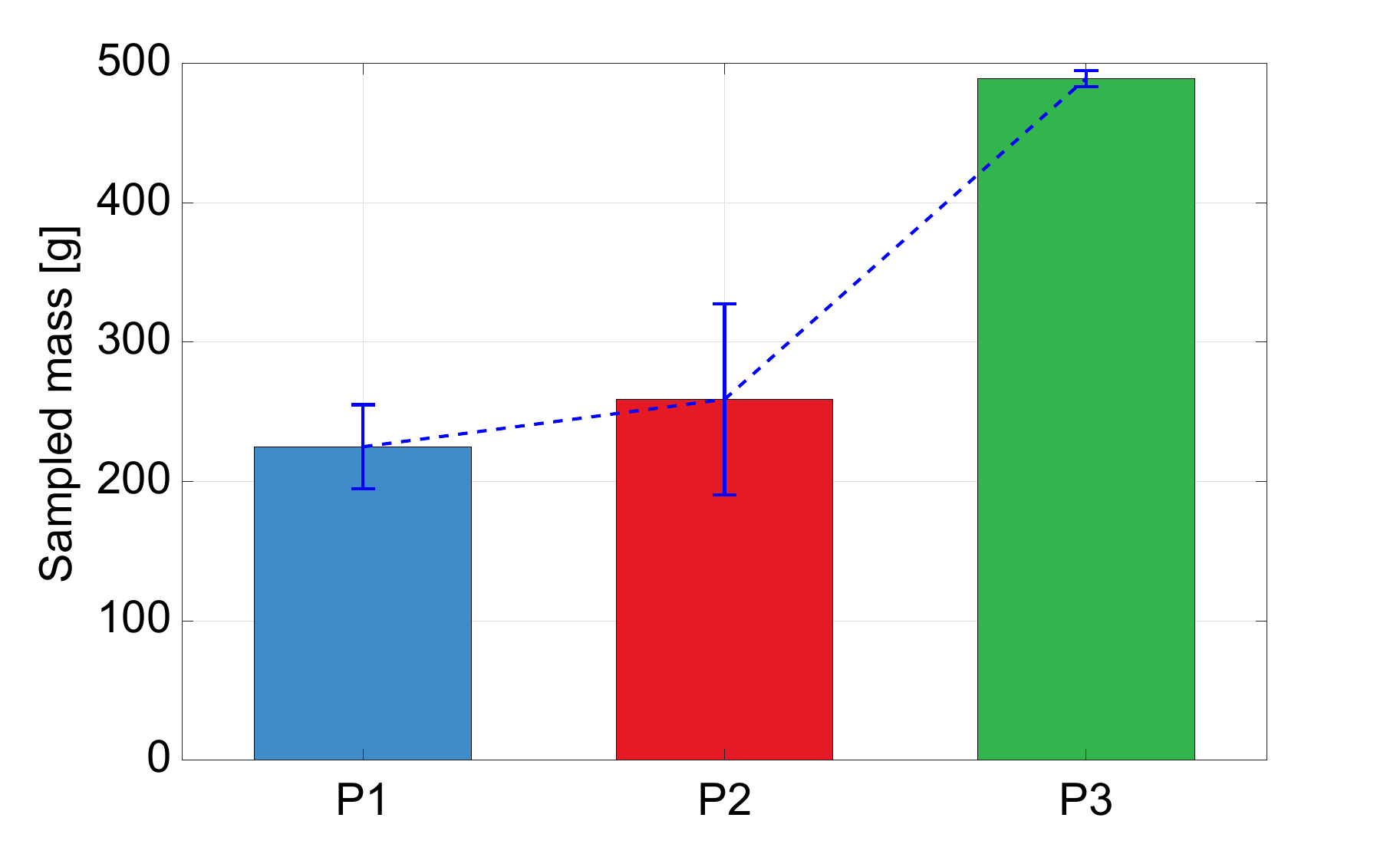} 
    \vspace{-1ex}
    \caption{Medium sand} 
    \label{mass_p2} 
  \end{subfigure} 
  \begin{subfigure}[b]{\linewidth}
    \centering
    \includegraphics[width=0.85\linewidth]{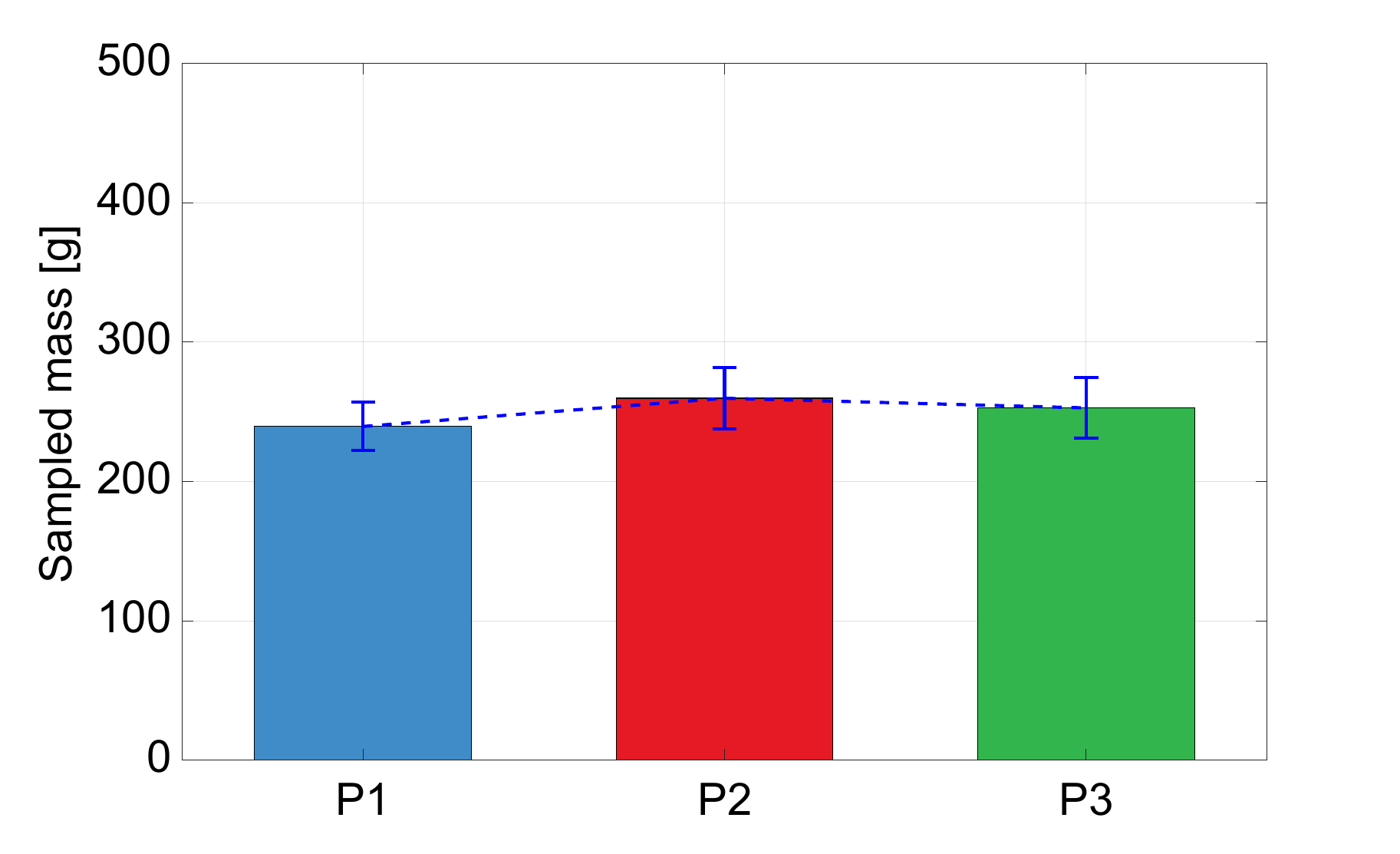} 
    \vspace{-1ex}
    \caption{Silt} 
    \label{mass_p3} 
  \end{subfigure}
\caption{Mean and standard deviation of each mass sampled by three patterns. The sampled masses of coarse sand and medium sand vary considerably with the pattern. No significant differences were observed as a function of coring patterns for silty sediment.}
\label{mass data} 
\end{figure}

\begin{figure}[tbp] 
\centering
  \begin{subfigure}[b]{\linewidth}
    \centering
    \includegraphics[width=0.8\linewidth]{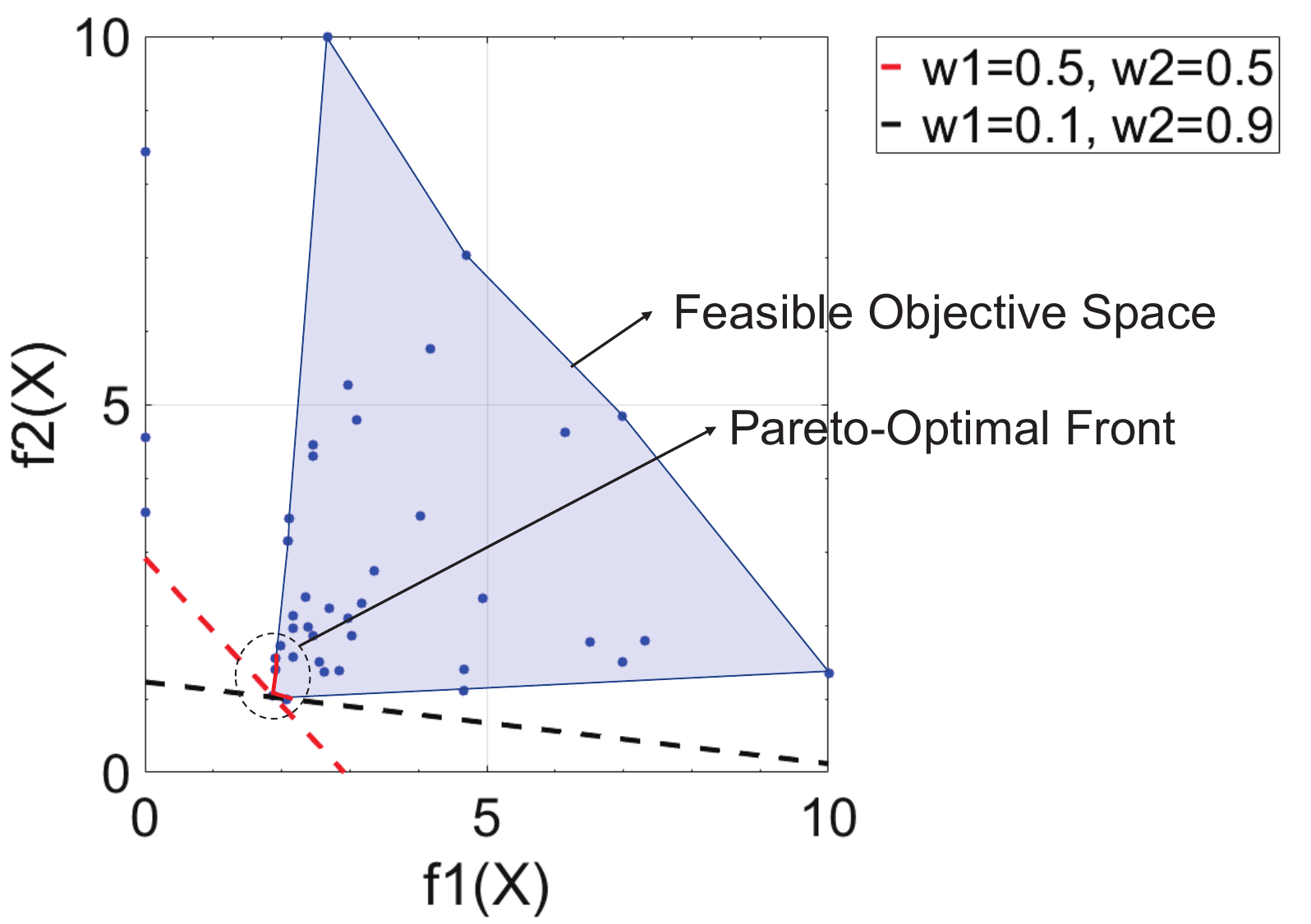} 
    \caption{Coarse sand} 
    \label{pareto_cs} 

  \end{subfigure}
  \begin{subfigure}[b]{\linewidth}
    \centering
    \includegraphics[width=0.8\linewidth]{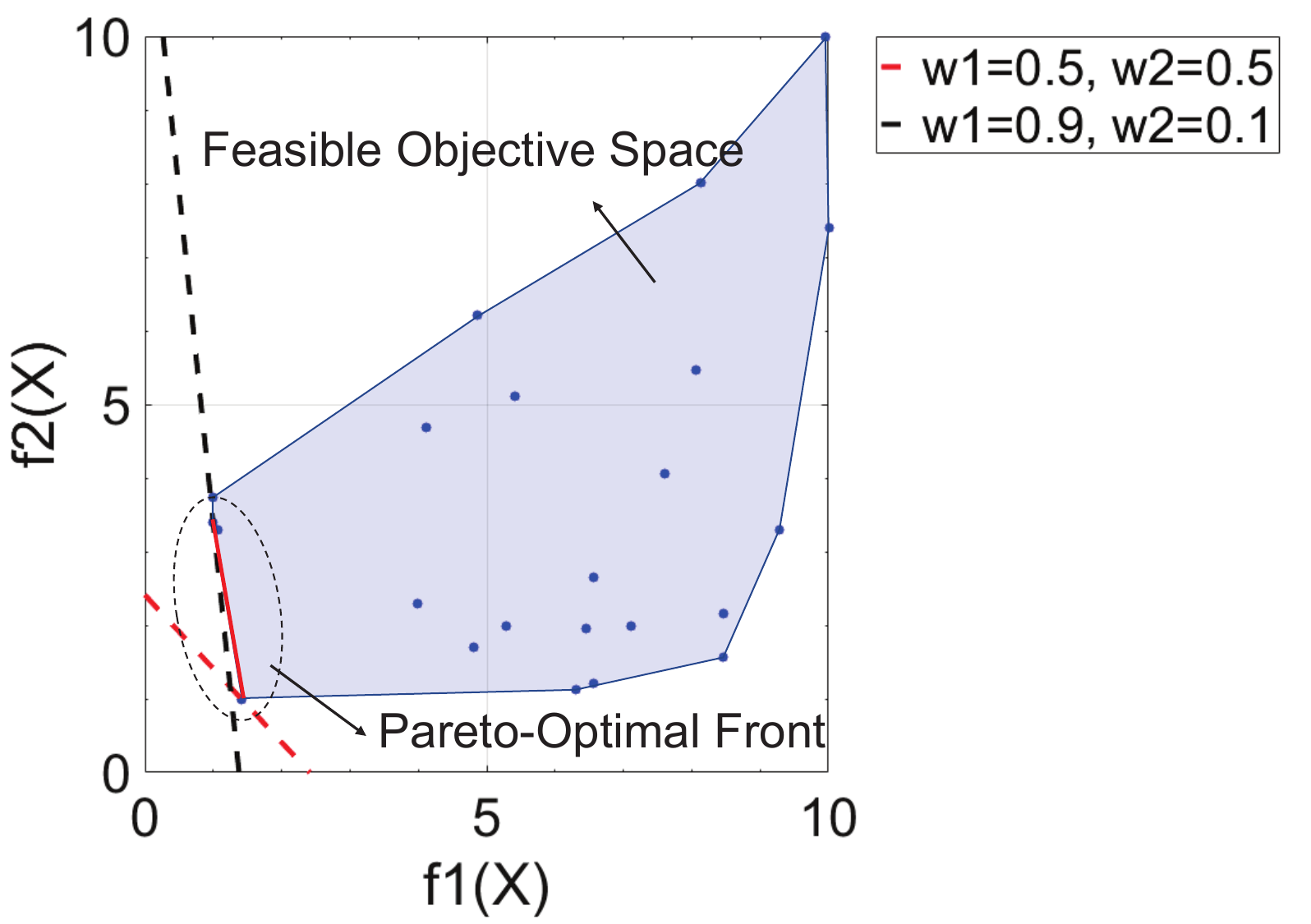} 
    \caption{Medium sand} 
    \label{pareto_ms} 
  \end{subfigure} 
  \begin{subfigure}[b]{\linewidth}
    \centering
    \includegraphics[width=0.8\linewidth]{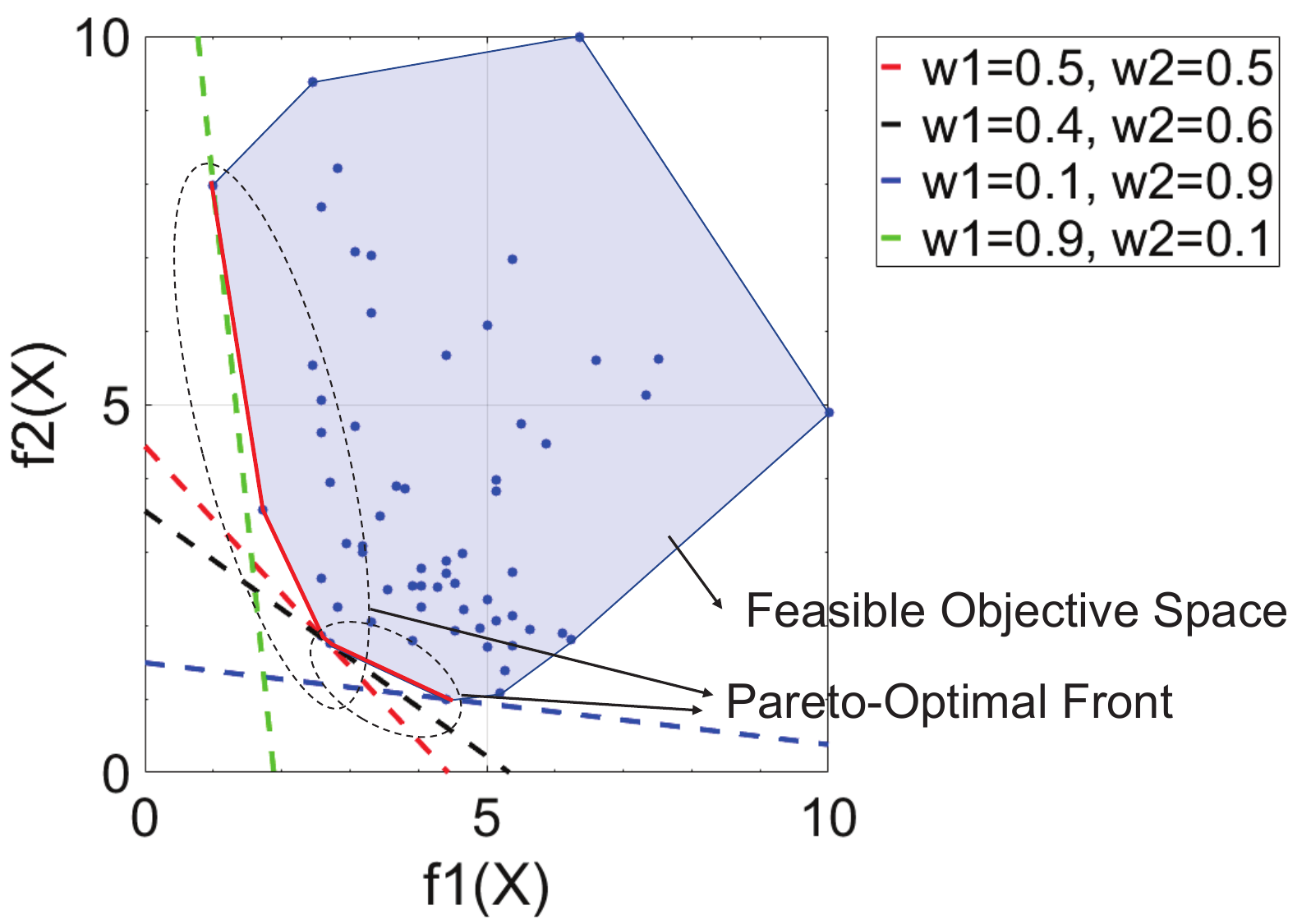} 
    \caption{Silt} 
    \label{pareto_fs} 
  \end{subfigure}
\caption{Pareto-optimal front (red line) in the feasible objective space (shaded area) and the weighted objective functions (dotted line) of (a) Coarse sand, (b) Medium sand, (c) Silt. The coarse and medium sand have two optimal patterns, respectively. The silt has four optimal patterns.}
\label{pareto-weighted} 
\end{figure}

\section{Data Analysis}
\label{sec:analysis}
Steps 1 and 2 measured four parameters: the mass of sampled sediment, penetration depth, penetration force, and motor currents. We analyzed the measured data to find the optimal values and characterize the sediments.

\subsection{Mass of Sampled Sediment}
The mean and standard deviation of the mass of sampled sediment based on each pattern shows the variation of the mass sampled by the pattern as shown in Fig. \ref{mass data}. The blue bar indicates the average of the sampled mass with $P_{1}$, the red bar corresponds to $P_{2}$, and the green bar to $P_{3}$.

In the case of coarse sand (Fig. \ref{mass_p1}), the amount of sampled mass in $P_{2}$ increased by $61 \%$ as compared to that in $P_{1}$. The amount of sampled mass in $P_{3}$ increased by $355 \%$ and $178 \%$ as compared to that in $P_{1}$ and $P_{2}$, respectively. In the case of medium sand (Fig. \ref{mass_p2}), the amount of sampled mass in $P_{2}$ increased by $15 \%$ as compared to that in $P_{1}$. The amount of sampled mass in $P_{3}$ increased by $117 \%$ and $88 \%$ as compared to that in $P_{1}$ and $P_{2}$, respectively. In the case of silt (Fig. \ref{mass_p3}), the amount sampled mass in $P_{2}$ increased by $8 \%$ as compared to that in $P_{1}$. The amount of sampled mass in $P_{3}$ increased by $5 \%$ as compared to that in $P_{1}$, but decreased by $2 \%$ as compared to that in $P_{2}$. 

Based on Fig. \ref{mass_p1}, Fig. \ref{mass_p2}, and Fig. \ref{mass_p3}, the sampled masses of coarse sand and medium sand increase as the pattern changes from $P_{1}$ to $P_{3}$ sequentially. In other words, a larger amount of coarse sand and medium sand was collected by $P_{3}$. In the case of silt, there was no significant difference between the patterns.

\subsection{Multiple Objective Optimization}
We posed the problem as a multiple objective optimization problem and defined the objective function using the weighted-sum method from Section \ref{sec:methodology}. The weighted-sum method addressed the set of objectives as one single objective by multiplying each objective by a user-defined weight \cite{marler_weighted_2010}. The feasible objective space was based on the $f_{1}(\mathbb{X})$ and $f_{2}(\mathbb{X})$ data sets for each sediment while solutions lie on a line called the \textit{Pareto-optimal front} \cite{lavin2015pareto}. As shown in Fig. \ref{pareto-weighted}, the shaded area is the feasible objective space and the bold red line is the Pareto-optimal front. 

\begin{figure*}[t]
 \centering 
  \includegraphics[width=0.75\linewidth]{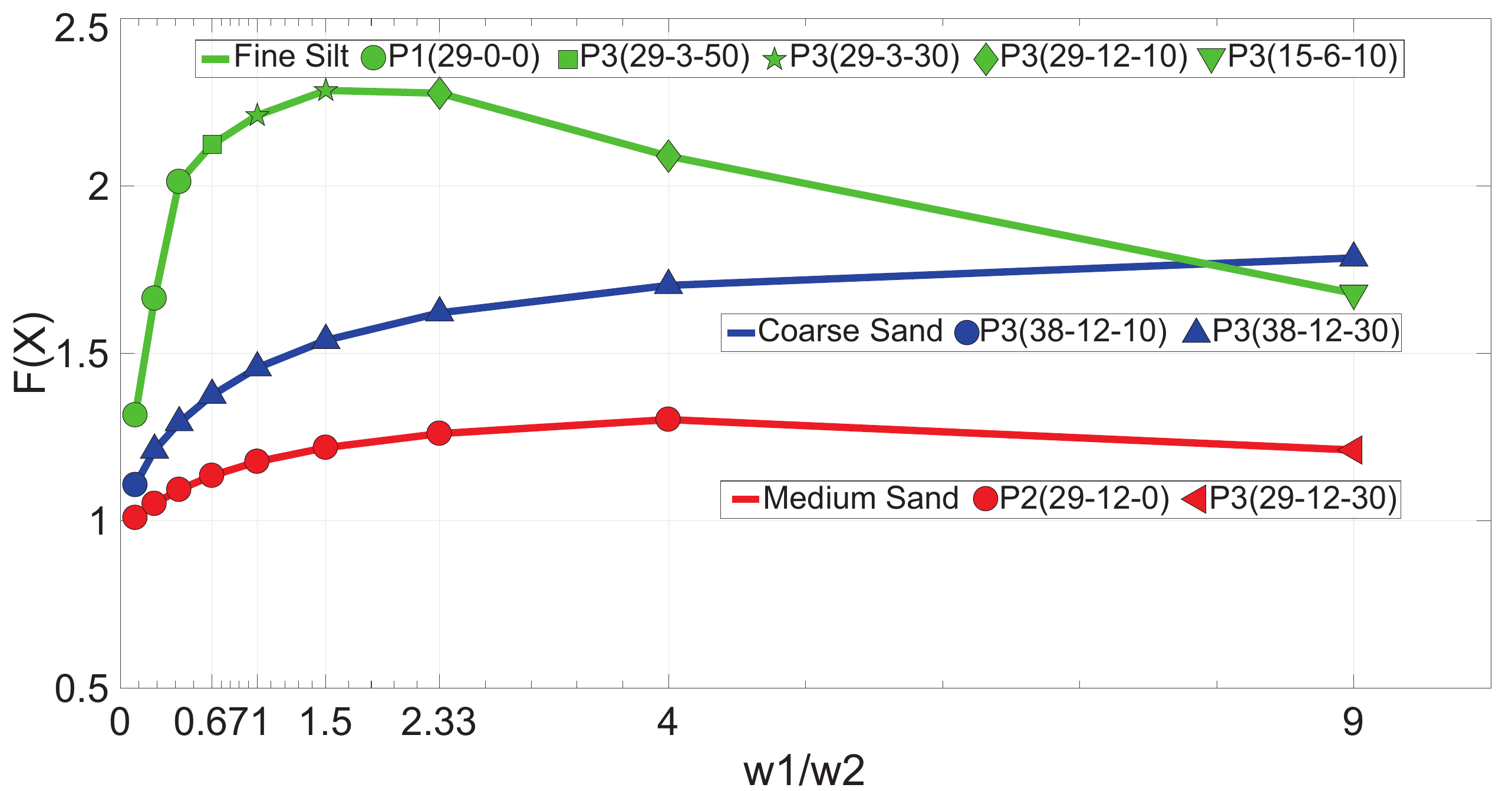} 
  \vspace{2ex}
  \caption{Diagram of the multiple objective optimization results where x-axis is the weight ratio between $w_{1}$ and $w_{2}$, and y-axis is the sum of the $f_{1}(\mathbb{X})$ and $f_{2}(\mathbb{X})$ multiplied by each weight; $F(\mathbb{X})=w_{1}f_{1}(\mathbb{X})+w_{2}f_{2}(\mathbb{X})$. In the case of coarse sand (blue line) and medium sand (red line), an optimal pattern is skewed despite the weight configuration. In the case of silt (green line), the optimal pattern varies depending on the weight. } 
  \label{mo_result} 
\end{figure*}

\begin{table}
\centering
  \vspace{0.03cm}
\caption {Results of the weighted-sum method: (a) Coarse sand, (b) Medium sand, (c) Silt.}
\label{Table6}

\begin{subtable}{\linewidth}
\begin{center}
\caption {Coarse sand}
\vspace{-1ex}
\label{Table6a}
\begin{tabular}{|c|c|c|c|c|c|c|c|c|} \hline
\multicolumn{2}{|c|}{Weight} & \multicolumn{3}{c|}{Objective Function} & \multicolumn{4}{c|}{Pattern} \\ \hline
$w_{1}$ & $w_{2}$ & $f_{1}(\mathbb{X})$ & $f_{2}(\mathbb{X})$ & $F(\mathbb{X})$ & $P$ & {$M_{1}$} & {$M_{2}$} & $f_{r}$ \\ \hline
\hline
0.1 & 0.9 & 2.07 & 1.00 & 1.107 & $P_{3}$ & 38 & 12 & 10   \\ \hline
0.2 & 0.8 & 1.87 & 1.05 & 1.211 & $P_{3}$ & 38 & 12 & 30   \\ \hline
0.3 & 0.7 & 1.87 & 1.05 & 1.293 & $P_{3}$ & 38 & 12 & 30   \\ \hline
0.4 & 0.6 & 1.87 & 1.05 & 1.375 & $P_{3}$ & 38 & 12 & 30   \\ \hline
0.5 & 0.5 & 1.87 & 1.05 & 1.457 & $P_{3}$ & 38 & 12 & 30   \\ \hline
0.6 & 0.4 & 1.87 & 1.05 & 1.539 & $P_{3}$ & 38 & 12 & 30   \\ \hline
0.7 & 0.3 & 1.87 & 1.05 & 1.621 & $P_{3}$ & 38 & 12 & 30   \\ \hline
0.8 & 0.2 & 1.87 & 1.05 & 1.703 & $P_{3}$ & 38 & 12 & 30   \\ \hline
0.9 & 0.1 & 1.87 & 1.05 & 1.785 & $P_{3}$ & 38 & 12 & 30   \\ \hline
\end{tabular}
\end{center}
\end{subtable} 

\begin{subtable}{\linewidth}
\begin{center}
\vspace{3ex}
\caption {Medium Sand}
\vspace{-1ex}
\label{Table6b}
\begin{tabular}{|c|c|c|c|c|c|c|c|c|} \hline
\multicolumn{2}{|c|}{Weight} & \multicolumn{3}{c|}{Objective Function} & \multicolumn{4}{c|}{Pattern} \\ \hline
$w_{1}$ & $w_{2}$ & $f_{1}(\mathbb{X})$ & $f_{2}(\mathbb{X})$ & $F(\mathbb{X})$ & $P$ & {$M_{1}$} & {$M_{2}$} & $f_{r}$ \\ \hline
\hline
0.1 & 0.9 & 1.41 & 1.00 & 1.041 & $P_{2}$ & 29 & 12 & 0   \\ \hline
0.2 & 0.8 & 1.41 & 1.00 & 1.082 & $P_{2}$ & 29 & 12 & 0   \\ \hline
0.3 & 0.7 & 1.41 & 1.00 & 1.123 & $P_{2}$ & 29 & 12 & 0   \\ \hline
0.4 & 0.6 & 1.41 & 1.00 & 1.164 & $P_{2}$ & 29 & 12 & 0   \\ \hline
0.5 & 0.5 & 1.41 & 1.00 & 1.205 & $P_{2}$ & 29 & 12 & 0   \\ \hline
0.6 & 0.4 & 1.41 & 1.00 & 1.246 & $P_{2}$ & 29 & 12 & 0   \\ \hline
0.7 & 0.3 & 1.41 & 1.00 & 1.287 & $P_{2}$ & 29 & 12 & 0   \\ \hline
0.8 & 0.2 & 1.41 & 1.00 & 1.328 & $P_{2}$ & 29 & 12 & 0   \\ \hline
0.9 & 0.1 & 1.00 & 3.39 & 1.239 & $P_{3}$ & 29 & 12 & 30   \\ \hline
\end{tabular}
\end{center}
\end{subtable} 

\begin{subtable}{\linewidth}
\begin{center}
\vspace{3ex}
\caption {Silt}
\vspace{-1ex}
\label{Table6c}
\begin{tabular}{|c|c|c|c|c|c|c|c|c|} \hline
\multicolumn{2}{|c|}{Weight} & \multicolumn{3}{c|}{Objective Function} & \multicolumn{4}{c|}{Pattern} \\ \hline
$w_{1}$ & $w_{2}$ & $f_{1}(\mathbb{X})$ & $f_{2}(\mathbb{X})$ & $F(\mathbb{X})$ & $P$ & {$M_{1}$} & {$M_{2}$} & $f_{r}$ \\ \hline
\hline
0.1 & 0.9 & 4.41 & 1.00 & 1.341 & $P_{1}$ & 29 & 0 & 0   \\ \hline
0.2 & 0.8 & 4.41 & 1.00 & 1.682 & $P_{1}$ & 29 & 0 & 0   \\ \hline
0.3 & 0.7 & 4.41 & 1.00 & 2.023 & $P_{1}$ & 29 & 0 & 0   \\ \hline
0.4 & 0.6 & 2.70 & 1.75 & 2.132 & $P_{3}$ & 29 & 3 & 50   \\ \hline
0.5 & 0.5 & 2.58 & 1.85 & 2.217 & $P_{3}$ & 29 & 3 & 30   \\ \hline
0.6 & 0.4 & 2.58 & 1.85 & 2.290 & $P_{3}$ & 29 & 3 & 30   \\ \hline
0.7 & 0.3 & 1.73 & 3.57 & 2.282 & $P_{3}$ & 29 & 12 & 10   \\ \hline
0.8 & 0.2 & 1.73 & 3.57 & 2.098 & $P_{3}$ & 29 & 12 & 10   \\ \hline
0.9 & 0.1 & 1.00 & 7.97 & 1.697 & $P_{3}$ & 15 & 6 & 10   \\ \hline
\end{tabular}
\end{center}
\end{subtable} 
\end{table}

Results of the multiple objective optimization via the weighted-sum method are shown in Fig. \ref{mo_result} and Table \ref{Table6}. In the case of coarse sand (Table \ref{Table6a}), an optimal pattern is skewed despite the weight configuration. Pattern $P_{3}$:~$38/12/30$ is the optimal pattern when the user has equal concern for mass recovery and power expended ($w_{1}=0.5$, $w_{2}=0.5$), mass-oriented sampling (when $w_{1}=0.6$ or higher / $w_{2}=0.4$ or less), or power-oriented sampling up to $w_{1}=0.2$, $w_{2}=0.8$. When the user is targeting a power-oriented sampling with $w_{2}=0.9$, pattern $P_{3}$:~$38/12/10$ is optimal. 

In the case of medium sand (Table \ref{Table6b}), an optimal pattern is also skewed regardless of the weights placed on either mass or power. Pattern $P_{3}$:~$29/12/30$ is the optimal pattern when the user is targeting mass-oriented sampling with $w_{1}=0.9$. Pattern $P_{2}$:~$29/12/0$ is the optimal pattern when the user is targeting mass-oriented sampling up to $w_{1}=0.8$, $w_{2}=0.2$, balanced sampling ($w_{1}=0.5$, $w_{2}=0.5$), or power-oriented sampling (when $w_{1}=0.4$ or less / $w_{2}=0.6$ or higher). 

In the case of silt (Table \ref{Table6c}), the optimal pattern varies depending on the weight. Pattern $P_{3}$:~$15/6/10$ is the optimal pattern when the user is targeting mass-oriented sampling with $w_{1}=0.9$. When the user is targeting mass-oriented sampling with $w_{1}=0.8$ and $w_{1}=0.7$, pattern $P_{3}$:~$29/12/10$ is the optimal pattern. When the user is targeting balanced sampling ($w_{1}=0.5$, $w_{2}=0.5$) or mass-oriented sampling with $w_{1}=0.6$, pattern $P_{3}$:~$29/3/30$ is the optimal pattern. When the user is targeting power-oriented sampling with $w_{2}=0.6$, pattern $P_{3}$:~$29/3/50$ is the optimal pattern. When the user is targeting power-oriented sampling (when $w_{1}=0.3$ or less / $w_{2}=0.7$ or higher), pattern $P_{1}$:~$29/0/0$ is optimal.

\subsection{Analysis on Sediment Disturbance}
Minimizing the sediment disturbance in core sampling is a challenging problem. Unavoidable disturbances occur during the whole process of sampling: drilling, recovery, transportation, handling, and early stages of analysis \cite{dai_sampling_2014}. Disturbances during the sampling process can cause disruption to the physical, geochemical, and biological condition of the sediment sample \cite{dai_sampling_2014} \cite{mogg_influence_2017} due to the various factors such as friction between the core liner and sediment, contamination by air exposure, and human error. Also, we realized that post-lab processes are necessary to measure parameters that can determine the disturbance of the sediment such as oxygen microprofiles, benthic oxygen flux rates, sediment solid phase analyses (chlorophyll a, organic carbon, and porosity), pore pressure, and secondary hydrate formation \cite{dai_sampling_2014} \cite{mogg_influence_2017}. 
We tested our sampling platform to characterize the relative amount of disturbance using the three patterns for driving the core liner into the sediment. In addition, we conducted manual sampling (i.e., hammer coring) to compare its result with the platform-based sampling. We measured disturbance as the total depth of sediment recovered and via visual observation of sampled sediment.

\begin{figure*}[t!] 
\centering
  \begin{subfigure}[b]{0.225\linewidth}
    \centering
    \includegraphics[width=0.98\linewidth]{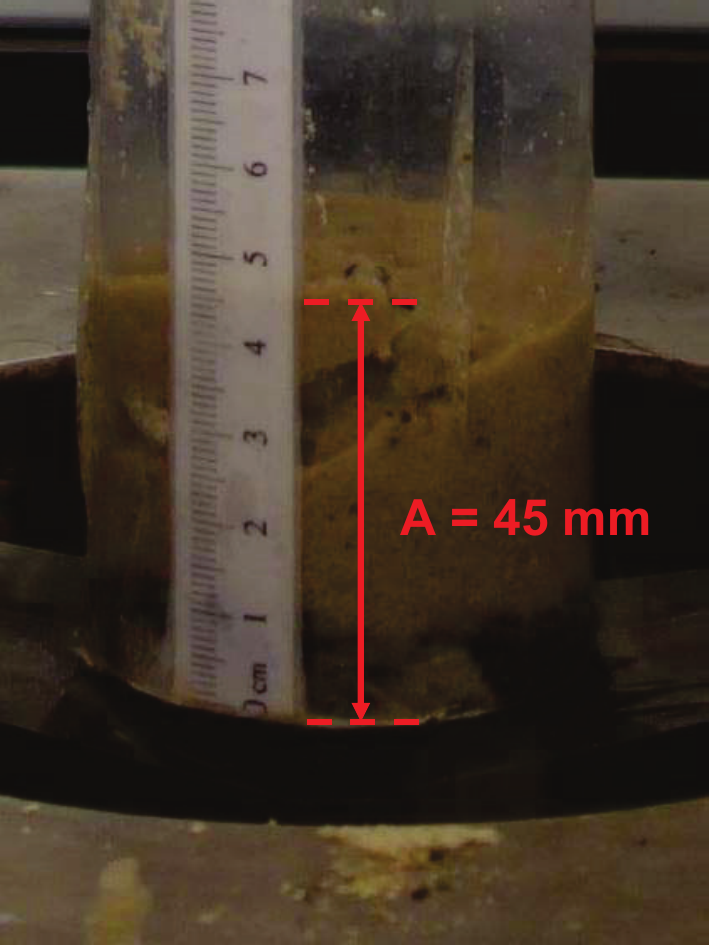} 
    \caption{Linear} 
    \label{p1dis}
  \end{subfigure} 
  \begin{subfigure}[b]{0.225\linewidth}
    \centering
    \includegraphics[width=0.98\linewidth]{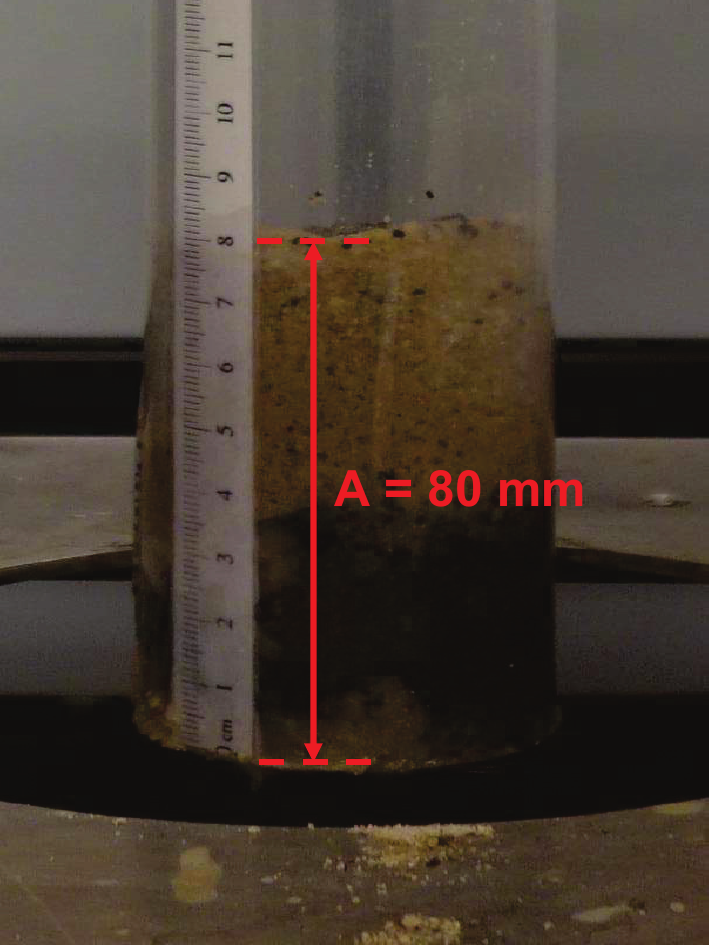} 
    \caption{Helix}
    \label{p2dis}
  \end{subfigure}
  \begin{subfigure}[b]{0.224\linewidth}
    \centering
    \includegraphics[width=0.98\linewidth]{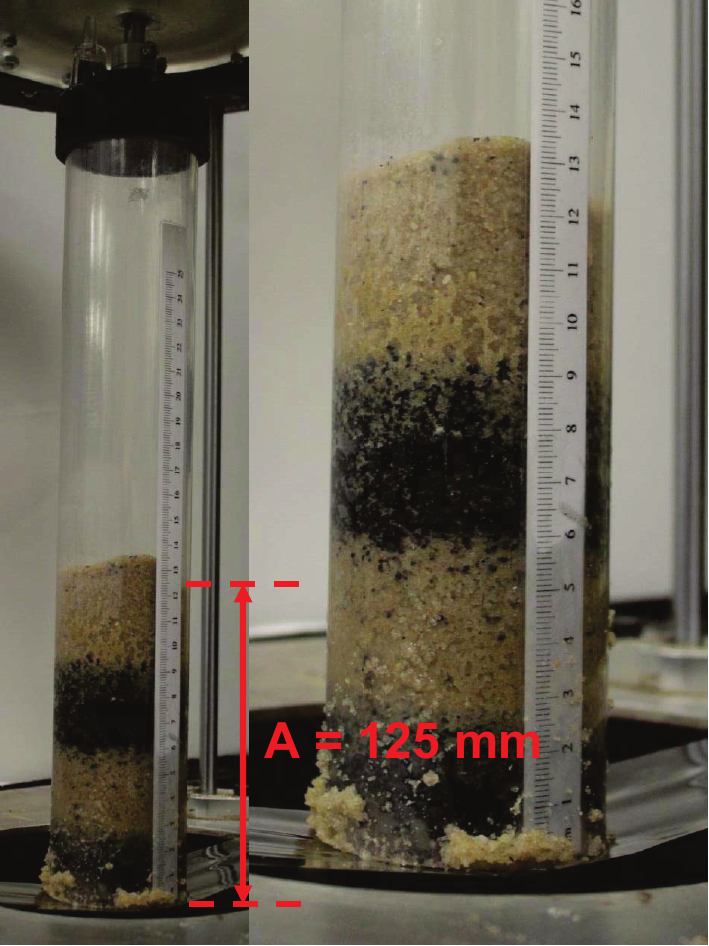} 
    \caption{Zig-zag}  
    \label{p3dis}
  \end{subfigure}
  \begin{subfigure}[b]{0.224\linewidth}
    \centering
    \includegraphics[width=0.98\linewidth]{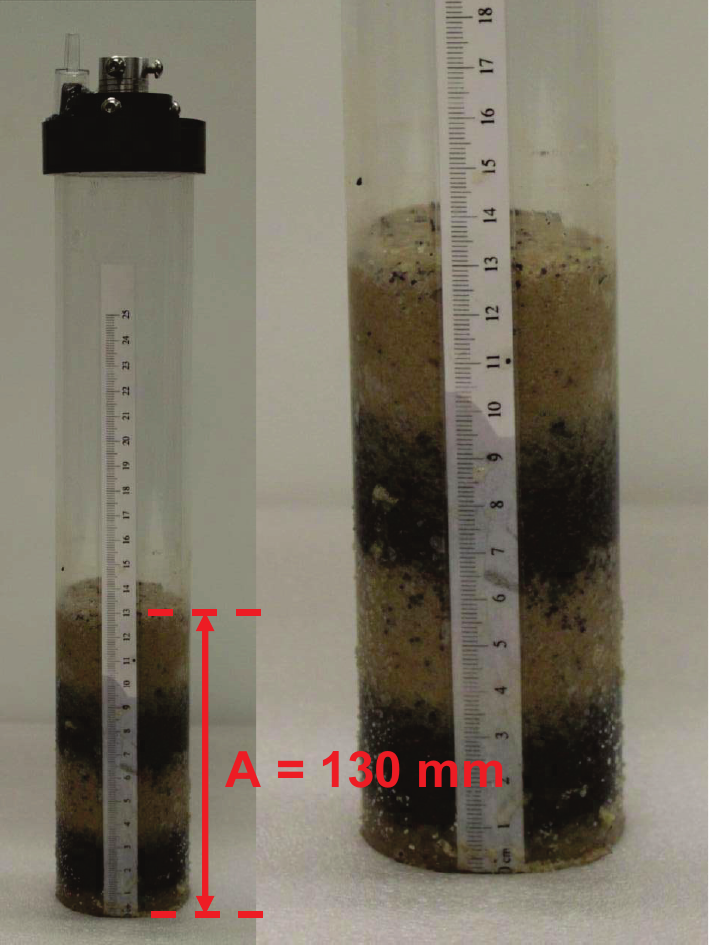} 
    \caption{Manual sampling}  
    \label{manualdis}
  \end{subfigure}
\caption{Results of the sediment sampling disturbance experiments to validate the optimal sampling patterns for the quality of the sediment samples. We measured sediment disturbance via visual observation of the deformation of the layers of sediments and the total depth of sediment recovered (A: core's length, and B: penetration depth of 200 mm); (a) $P_{1}$: Linear (A/B $=0.225$), (b) $P_{2}$: Helix (A/B $=0.4$), (c) $P_{3}$: Zig-zag (A/B $=0.625$), and (d) Manual sampling by hammer coring (A/B $=0.65$).}
\label{sediment disturbance} 
\end{figure*}

For visual observation and analysis of sampled sediments, we used a coarse-grained sand layered with distinct red and tan colors (7 layers total and each layer was 30 mm thick). Three optimal patterns (linear ($P_{1}$:~$38/0/0$), helix ($P_{3}$:~$38/12/0$), and zig-zag ($P_{3}$:~$38/12/10$)) that we found in Section \ref{sec:experiment} have applied. As shown in the figures, the recovery ratio (A/B)\footnote[3]{Ratio A/B where “A” is the distance from the top of the sediment core to the bottom (core's length), and “B” is the penetration depth that we set as 200 mm.} was greatest with $P_{3}$: Zig-zag (A/B $=0.625$, Fig. \ref{p3dis}), followed by $P_{2}$: Helix (A/B $=0.4$, Fig. \ref{p2dis}) and $P_{1}$: Linear (A/B $=0.225$, Fig. \ref{p1dis}), which was expected based on the previous experiment results in Section \ref{step1} and \ref{step2}. Moreover, from the layered structure (the layer of tan and red sediment) of the sample taken with the three patterns based on the automated system and the manually sampled sample could not be concluded that there was a difference only by visual observation. For more precise analysis on the sediment disturbance, additional post-lab processing is required. We leave this for future work.

\subsection{Summary}
Analysis of experimental results shows that the optimal pattern changed depending on the sediment pattern, sediment type, and user’s objectives. 
In the case of coarse sand, pattern $P_{3}$ (zig-zag) performed best, regardless of objective preference, with greater linear velocity $v$ and angular velocity $\omega_{r}$ improving both mass collection and power efficiency. The frequency of oscillation $f_{r}$ did not strongly impact mass collection or power efficiency.

With medium sand, pattern $P_{2}$ (helical) is best unless mass collection is strongly preferred ($w_{1}>0.9$) over efficiency. In that case, $P_{3}$ is better. Using $P_{2}$, the faster angular velocity $\omega_{r}$ demonstrated better performance. However, the fastest linear velocity did not assure the best result. Using $P_{3}$ (mass collection preference), higher frequencies tended to improve mass collection, but not by much.

In the case of silt, pattern $P_{3}$ is best unless power efficiency is moderately preferred. In that case, pattern $P_{1}$ (linear) is better. The interaction between $f_{1}(\mathbb{X})$ and $f_{2}(\mathbb{X})$ was not strong, however.

\section{Conclusions and Future Work}
\label{sec:conclusion}
In this paper, we developed a robotic sediment sampling platform to analyze the characteristics of sediment sampling methods depending on various types of sediment. We conducted experiments with three different sampling patterns (linear, helical, and zig-zag) and three sediment types (coarse sand, medium sand, and silt) and determined all three sampling patterns added value to the core sampler as no single pattern maximized performance in all cases. We found that the optimal sampling pattern varied with the type of sediment and could be optimized based on different sampling objectives. In general, the zig-zag pattern $P_{3}$ is useful for all three sediment types, but the linear pattern $P_{1}$ excels for silt when power efficiency is moderately preferred and the helical pattern $P_{2}$ excels for medium sand unless mass collection is strongly preferred.

While we expect the results of this research will serve as useful data for the development of more advanced sediment monitoring systems and other applications in the future, we acknowledge several limitations to our work. First of all, our experiments are limited to one coring tool and we did not explore the effects of diameter, cutting shoe angle, and wall thickness on coring performance. In addition, retention of the sample in the coring tool was a priority, but our solution lacks robustness. Rather than implementing an automated cap system, which is difficult to fabricate, we used a check-valve and manual blocking of the end, when necessary. This retained sediment in the core successfully, but impacts sample quality. More importantly, we did not investigate clay, which is rather common to rivers and lakes. This is due to a limitation of our immediate small-robot design with a focus on man-made reservoirs, in which silt and sand are more common.

During development and experimentation, we discovered numerous challenges and avenues for future work. Setting up the experimental environment was a laborious process. During the repeated experiments, re-setting the same test environment to avoid and minimize test environment bias was critical. Minimizing bias remains challenging in this research. In addition, we realized that the interaction between penetration force and depth, penetration force and time, and penetration depth and time is important, yet unaddressed. In this regard, we have collected additional experimental data for further analysis. Finally, this research is based on homogeneous sediment. Experiments based on multiple layers of different types of sediment should be considered to improve correspondence with natural environments.

\section*{Acknowledgment}
We thank the anonymous reviewers whose comments and suggestions helped improve and clarify this manuscript. This work was partially supported through the ``NSF Center for Robots and Sensors for the Human Well-Being, NSF Grant No. 1439717" and the Arequipa Nexus Institute. Assistance for soil characterization, provided by Danielle Winter in the Agricultural and Biological Engineering Dept., Purdue University, was greatly appreciated.

\bibliographystyle{IEEEtran}
\bibliography{main}

\begin{IEEEbiography}[{\includegraphics[width=1in,height=1.25in,clip,keepaspectratio]{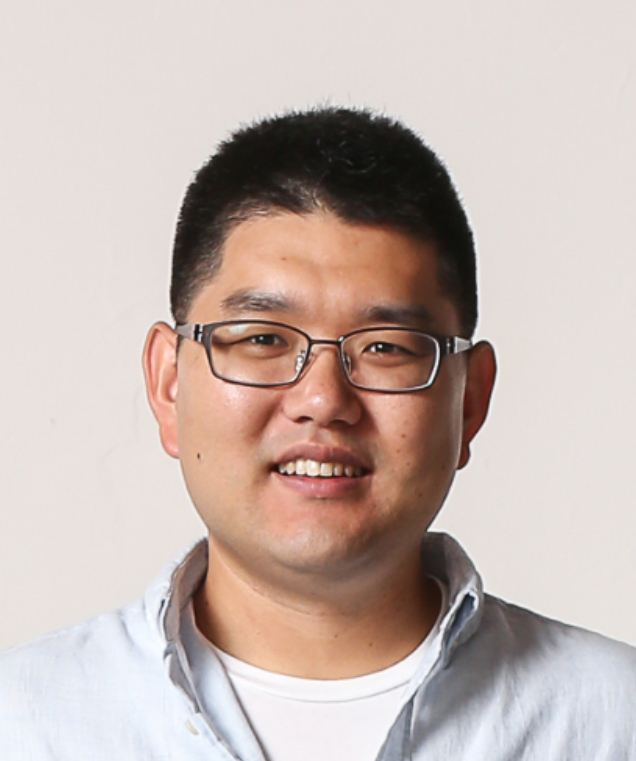}}]%
{Jun Han Bae} is a Ph.D. student in the Department of Computer and Information Technology at Purdue University (West Lafayette), under the advisement of Dr. Byung-Cheol Min and Dr. Richard M. Voyles. Prior to beginning the Ph.D.  program, Jun Han obtained his B.S. degree in Mechanical Engineering from Yonsei University, Republic of Korea in 2011, and M.S. in Mechanical Engineering Technology degree from Purdue University, West Lafayette, IN USA, in 2014. From 2012 to 2013, prior to his master's degree program, Jun Han held a junior manager position at STX offshore \& shipbuilding company.  His research interests are prototyping, robot design and control for an extreme environment, and mechanical design optimization. In addition to pursuing his Ph.D. degree, Jun Han developed an interest in robotic system applications in environmental monitoring and sampling. 
\end{IEEEbiography}
\vskip 0pt plus -1fil

\begin{IEEEbiography}[{\includegraphics[width=1in,height=1.25in,clip,keepaspectratio]{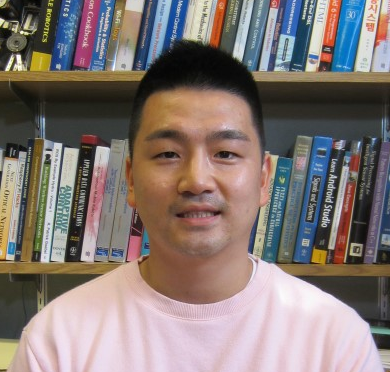}}]%
{Wonse Jo} received the B.S. degree in robotics engineering from Ho-Seo University, Asan, South Korea, in 2013, and the M.S. degree in electronics and radio engineering from Kyung Hee University, Yongin, South Korea, in 2015. Currently, he is pursuing a Ph.D. degree in technology with a specialization in human-multi robot interaction from Purdue University, West Lafayette, IN, USA.
\end{IEEEbiography}
\vskip 0pt plus -1fil

\begin{IEEEbiography}[{\includegraphics[width=1in,height=1.25in,clip,keepaspectratio]{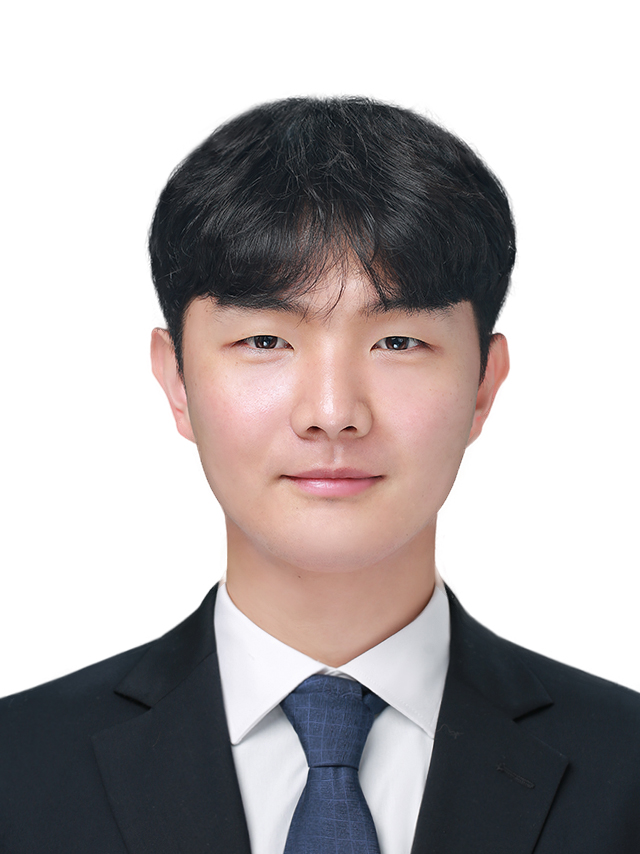}}]%
{Jee Hwan Park} received the B.S. degree in mechanical engineering from Purdue University, West Lafayette, USA, in 2018, and is currently enrolled in M.S. program in mechanical engineering from Purdue University, West Lafayette, USA. He is a graduate researcher at SMART Laboratory with Purdue University, West Lafayette, IN, USA.
\end{IEEEbiography}
\vskip 0pt plus -1fil

\begin{IEEEbiography}[{\includegraphics[width=1in,height=1.25in,clip,keepaspectratio]{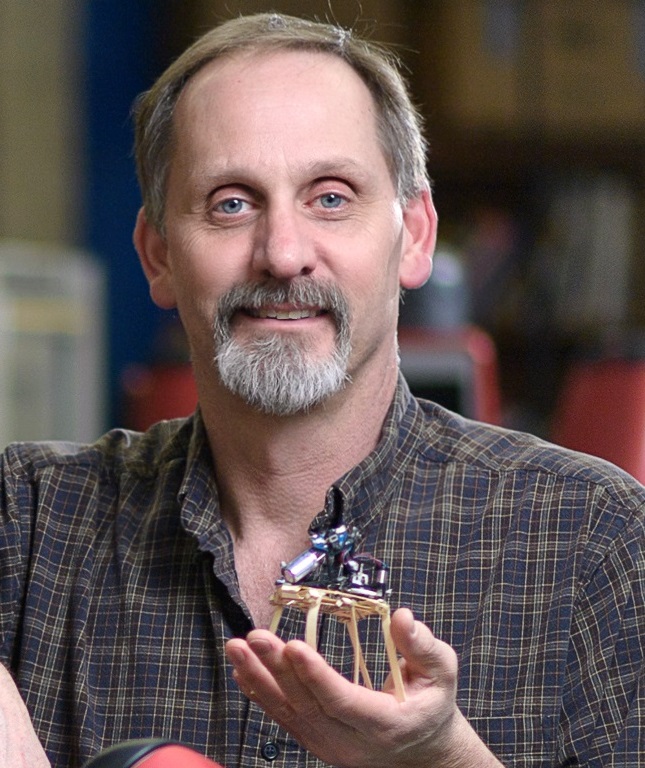}}]%
{Richard M. Voyles} (M’90–SM’00) received the B.S. degree in electrical engineering from Purdue University, West Lafayette, IN, USA, in 1983, the M.S. degree in manufacturing systems engineering from the Department of Mechanical Engineering, Stanford University, Stanford, CA, USA, in 1989, and the Ph.D. degree in robotics from the School of Computer Science, Carnegie Mellon University, Pittsburgh, PA, USA, in 1997.

He is the head of the Collaborative Robotics Lab, Director of the Purdue Robotics Accelerator and Daniel C. Lewis Professor of the School of Engineering Technology at Purdue University. His experience includes IBM, Integrated Systems, Inc., tenured positions at the University of Minnesota and University of Denver, Program Director at the National Science Foundation and Assistant Director of Robotics and Cyber-Physical Systems, Office of Science and Technology Policy, at the White House.

\end{IEEEbiography}
\vskip 0pt plus -1fil

\begin{IEEEbiography}[{\includegraphics[width=1in,height=1.25in,clip,keepaspectratio]{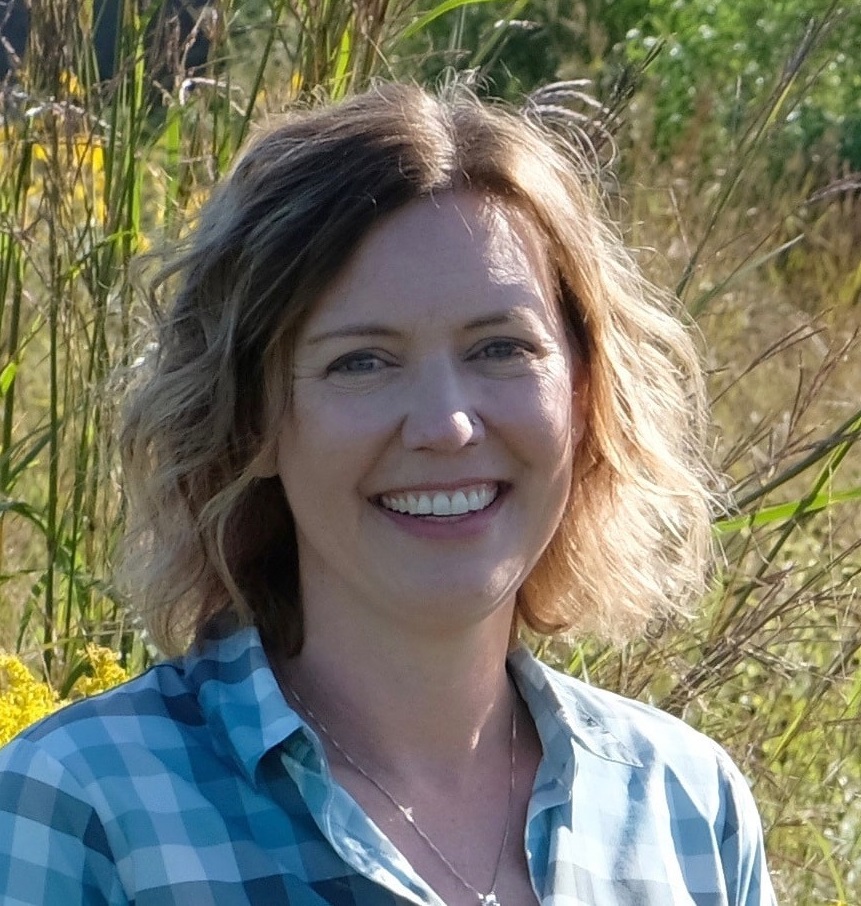}}]%
{Sara K. McMillan} is an Associate Professor in Agricultural and Biological Engineering at Purdue University. She is a recognized leader in water quality impacts of ecological restoration. Her research team employs an integrated approach of environmental sampling and sensor networks to monitor conditions, controlled experiments in the laboratory, and computer modeling. She received her Ph.D. in Environmental Science and Engineering from the University of North Carolina at Chapel Hill and BS in Civil \& Environmental Engineering from the University of Iowa. Prior to coming to Purdue, she spent 5 years as a professor at the University of North Carolina at Charlotte and several years working as a professional engineer on the impacts of changing land use and climate on water quality. 
\end{IEEEbiography}
\vskip 0pt plus -1fil

\begin{IEEEbiography}[{\includegraphics[width=1in,height=1.25in,clip,keepaspectratio]{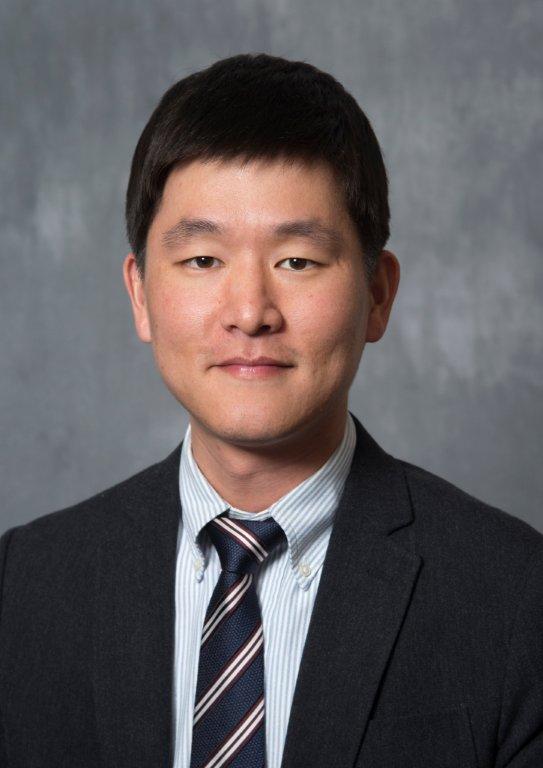}}]%
{Byung-Cheol Min} (M’14) received the B.S. degree in electronics engineering and the M.S. degree in electronics and radio engineering from Kyung Hee University, Yongin, South Korea, in 2008 and 2010, respectively, and the Ph.D. degree in technology with a specialization in robotics from Purdue University, West Lafayette, IN, USA, in 2014. 
 
He is an Assistant Professor of Department of Computer and Information Technology and the Director of the SMART Laboratory with Purdue University, West Lafayette, IN, USA. Prior to this position, he was a Postdoctoral Fellow with the Robotics Institute, Carnegie Mellon University, Pittsburgh, PA, USA. His research interests include multi-robot systems, human-robot interaction, robot design and control, with applications in field robotics and assistive technology and robotics.
 
He is a recipient of the NSF CAREER Award (2019), Purdue PPI Outstanding Faculty in Discovery Award (2019), and Purdue CIT Outstanding Graduate Mentor Award (2019).
\end{IEEEbiography}

\end{document}